\documentclass[11pt]{article}

\usepackage[preprint]{acl}

\usepackage{times}
\usepackage{latexsym}

\usepackage[T1]{fontenc}

\usepackage[utf8]{inputenc}

\usepackage{microtype}

\usepackage{inconsolata}

\usepackage{graphicx}

\usepackage{amsmath,amssymb,amsfonts}
\usepackage{subfig}
\usepackage{multirow}
\usepackage{booktabs}
\usepackage[ruled,linesnumbered]{algorithm2e}
\usepackage{amsthm}
\usepackage{makecell}

\title{Entropy-driven Dynamic Gradient Compression for Efficient LLM Training}

\author{
  \textbf{Qingao Yi\textsuperscript{1}},
  \textbf{Jiaang Duan\textsuperscript{2}},
  \textbf{Jun Zhang\textsuperscript{4}},
  \textbf{Haiyan Zhao\textsuperscript{1}},
  \textbf{Shiyou Qian\textsuperscript{2}},
  \textbf{Dingyu Yang\textsuperscript{3}},
\\
  \textbf{Jian Cao\textsuperscript{2}},
  \textbf{Jinghua Tang\textsuperscript{2}}
\\
\\
  \textsuperscript{1}University of Shanghai for Science and Technology, Shanghai, China,
\\
  \textsuperscript{2}Shanghai Jiao Tong University, Shanghai, China,
\\
  \textsuperscript{3}Zhejiang University, Hangzhou, China,
  \textsuperscript{4}Alibaba Group, Hangzhou, China
}

\begin{document}
\maketitle
\begin{abstract}
Training large language models (LLMs) at scale incurs substantial communication overhead, while static gradient compression cannot adapt to gradient evolution and may degrade model quality. We propose EDGC, an entropy-driven dynamic gradient compression framework that adapts compression ranks to gradient entropy during training. EDGC combines efficient entropy estimation through gradient sampling, a theoretical model relating entropy to compression rank under a bounded-error constraint, and window-based rank adjustment across pipeline stages. Experiments on 32-V100 and 64-H100 GPU clusters training GPT2 models with 2.5B and 12.1B parameters show that EDGC reduces communication latency by up to 46.45\% and end-to-end training time by 16.13\%, while maintaining model quality.
\end{abstract}

\section{Introduction}

The rapid growth of large language models (LLMs) has outpaced the compute and memory capacity of individual GPUs~\cite{1,43}, making distributed training essential. Modern systems commonly combine data parallelism (DP), pipeline parallelism (PP), and tensor parallelism (TP)~\cite{6,2,16}. Although this 3D parallelism enables large-scale training, the resulting inter-GPU communication can substantially limit performance, particularly across bandwidth-constrained nodes.

Existing approaches reduce communication through topology optimization~\cite{26}, synchronization and scheduling~\cite{29,32}, state partitioning~\cite{83}, or gradient compression~\cite{27}. Systems such as Megatron-LM~\cite{45} optimize communication across parallel dimensions, while Optimus-CC~\cite{35} compresses DP and pipeline traffic. Nevertheless, most compression methods use fixed, empirically selected rates. Such rates cannot follow gradient evolution: aggressive compression may harm early training, whereas conservative compression wastes bandwidth after gradients stabilize. Moreover, communication demand varies across pipeline stages.

Our measurements reveal three relevant properties of LLM gradients (\S\ref{sec:observation}): their entropy generally decreases during training, their values become increasingly concentrated near zero, and their matrices exhibit structural correlations. Together, these properties indicate that gradients become progressively more compressible and motivate adapting the rank of low-rank approximation~\cite{52} over time.

We therefore propose EDGC, an entropy-driven dynamic gradient compression framework (\S\ref{sec: Design}). EDGC comprises three components. First, a gradient data sampler estimates entropy from a small subset of gradients. Second, a compression quantification model relates entropy changes to compression rank under a bounded-error constraint. Third, a dynamic alignment compressor adjusts ranks by time window and pipeline stage, avoiding unstable early compression while balancing stage-specific communication.

We evaluate EDGC on a 32-GPU V100 cluster with 32-Gbps networking and a 64-GPU H100 cluster with 400-Gbps networking, training GPT2 models with 2.5B and 12.1B parameters, respectively. Against uncompressed Megatron-LM~\cite{45}, fixed-rank PowerSGD~\cite{52}, and Optimus-CC~\cite{35}, EDGC reduces communication latency by up to 45.8\% and 46.45\%, and end-to-end training time by up to 14.64\% and 16.13\%, on the two model scales while maintaining comparable model quality.

Our contributions are threefold: (1) we characterize the temporal evolution of LLM gradients and derive an entropy-based relationship between compression error and rank; (2) we design and implement a window- and stage-aware dynamic compression framework; and (3) we demonstrate consistent communication and training-time reductions across model scales, GPU generations, and network bandwidths.

\section{Background}

\paragraph{Parallel Training Strategies.}
\label{sec:3D Parallelism}

Large-scale LLM training commonly combines data, pipeline, and tensor parallelism as 3D parallelism. Because tensor parallelism requires frequent collective communication, it is typically confined to a node to exploit high-bandwidth NVLink, whereas data and pipeline parallelism may span nodes. Their inter-node traffic can consequently become a major training bottleneck.

\paragraph{Low-rank Gradient Compression.}
\label{sec:gradient-compression}

Low-rank compression approximates a gradient matrix \(G\) as \(G \approx PQ^\top\) and communicates the smaller factor matrices instead of \(G\)~\cite{70}. PowerSGD~\cite{52} efficiently obtains these factors through power iteration and orthogonalization. A lower rank reduces communication volume but generally increases approximation error, making rank the central control variable in EDGC.

\section{Observations and Insights}
\label{sec:observation}

We analyze gradient evolution during LLM training to guide the design of EDGC. We pre-train GPT2-345M~\cite{34} on OpenWebText~\cite{72} and BERT~\cite{42} on Wikipedia~\cite{85}. The eight-node cluster contains two 32-GB NVIDIA V100-SXM2 GPUs, 16 CPU cores, and 128 GB of memory per node. Nodes communicate over 32-Gbps Ethernet, while intra-node communication uses 300-Gbps NVLink.

\subsection{Observations}
Training an LLM requires ongoing gradient computation and the update of billions of weights. These weights create a dynamic system that transitions from disorder to order. Initially, they are randomly initialized, leading to high uncertainty characterized by high entropy. As training progresses and weights undergo multiple gradient updates, they stabilize, leading to gradients conveying similar information. The evolution of LLM gradients reveals three key insights: a decrease in gradient information entropy, zero-centralization of gradients, and correlation among gradients.

\emph{\textbf{Observation 1: Gradient information entropy transitions from an unstable to a dynamically stable state during training.}}

\textbf{Definition 1: Entropy of Continuous Random Variable.} \emph{For a continuous random variable \(X\) with probability density function \(f(x)\), the information entropy is defined as \(H(X)\), i.e.,}
\begin{align}
    H(X) = -\int_{-\infty}^{\infty} f(x) \log f(x) \, dx
    \label{eq:1}
\end{align}

\emph{Entropy \(H(X)\) quantifies the uncertainty associated with the outcomes of the random variable \(X\). Higher values indicate greater uncertainty.}

\begin{figure}[tbp]
    \centering
    \subfloat[GPT2-345M]{
    \includegraphics[width = 0.45\linewidth]{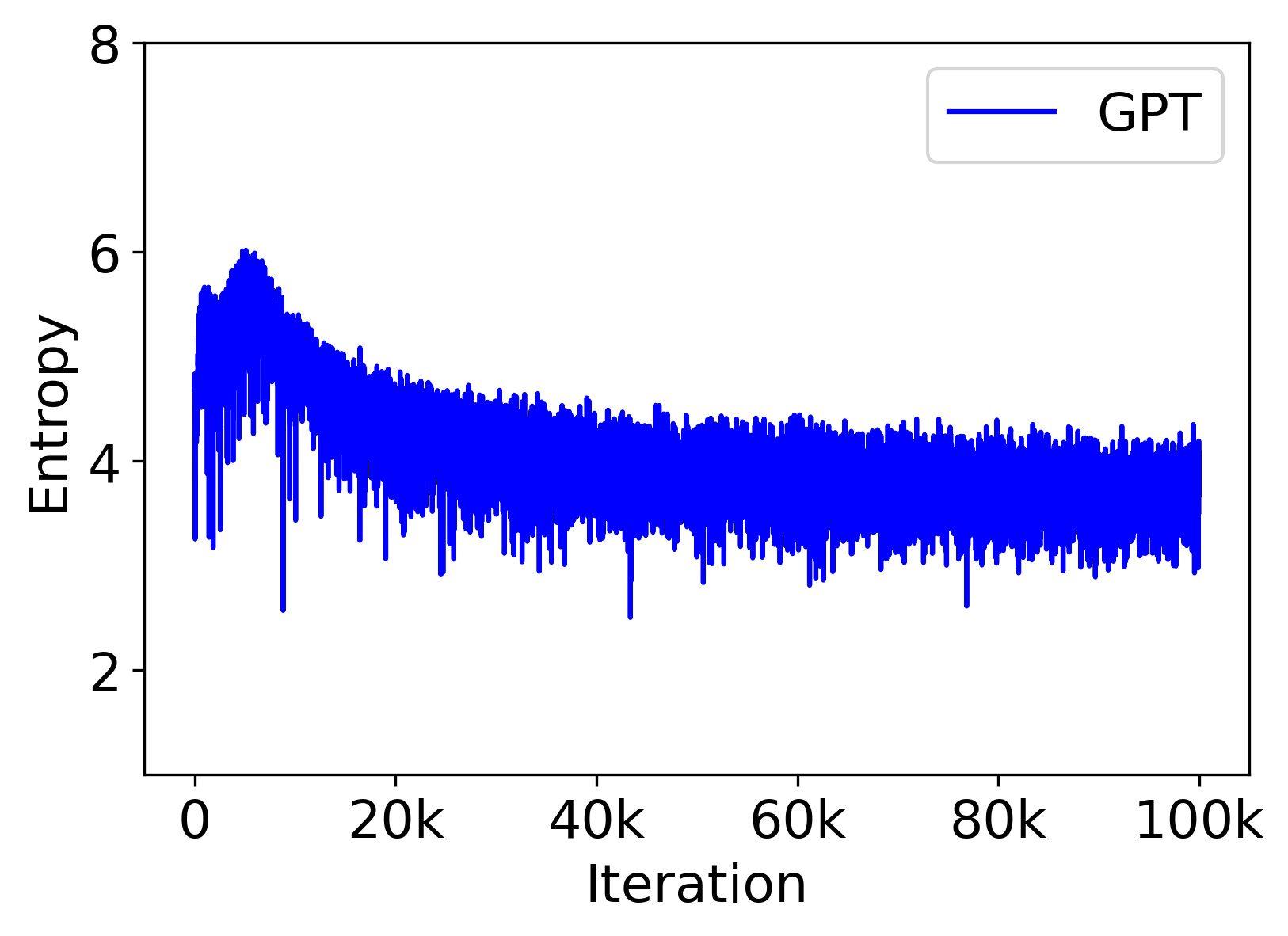}
    \label{fig:GPT}
    }
    \subfloat[BERT]{
    \includegraphics[width = 0.45\linewidth]{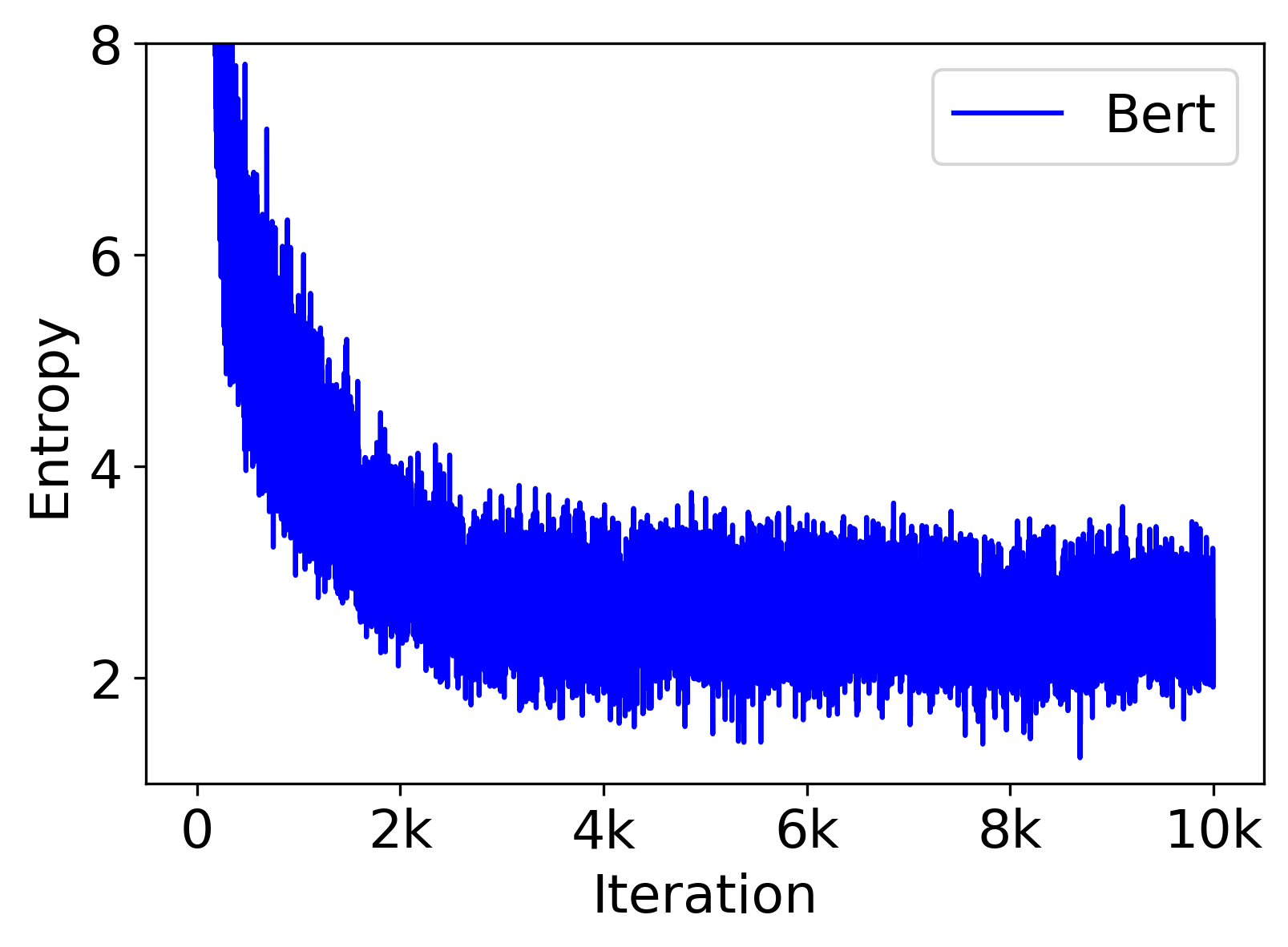}
    \label{fig:Bert}
    }
    \vspace{-2mm}
\caption{Entropy changes during iteration.}
\vspace{-5mm}
\label{fig:entropy_over_iteration}
\end{figure}

Figure \ref{fig:entropy_over_iteration} illustrates the entropy evolution for GPT2-345M and BERT. Both models exhibit an initial high-entropy phase followed by stabilization, yet they differ significantly in two dimensions due to their distinct architectures and training dynamics.
First, the duration of instability in LLMs varies significantly. For example, GPT2 and BERT achieve stability around iteration 100,000 and 10,000, respectively. This underscores the need for tailored compression strategies: BERT can be compressed more rapidly due to its earlier stability, while GPT2 necessitates a slower, more cautious approach.
Second, entropy fluctuations exhibit varying magnitudes during the convergence process. For GPT2, the entropy remains stable within a narrow range of 3.3--4.3, while BERT's entropy shows slightly greater variability at 1.8-3.8.
Consequently, precise adjustments to compression rates are necessary for different LLMs to ensure stability.

\emph{\textbf{Observation 2: As iterations progress, the gradient variation range narrows, indicating a trend toward zero-centralization.}}

\begin{figure}[tbp]
    \centering
    \includegraphics[width=0.48\linewidth]{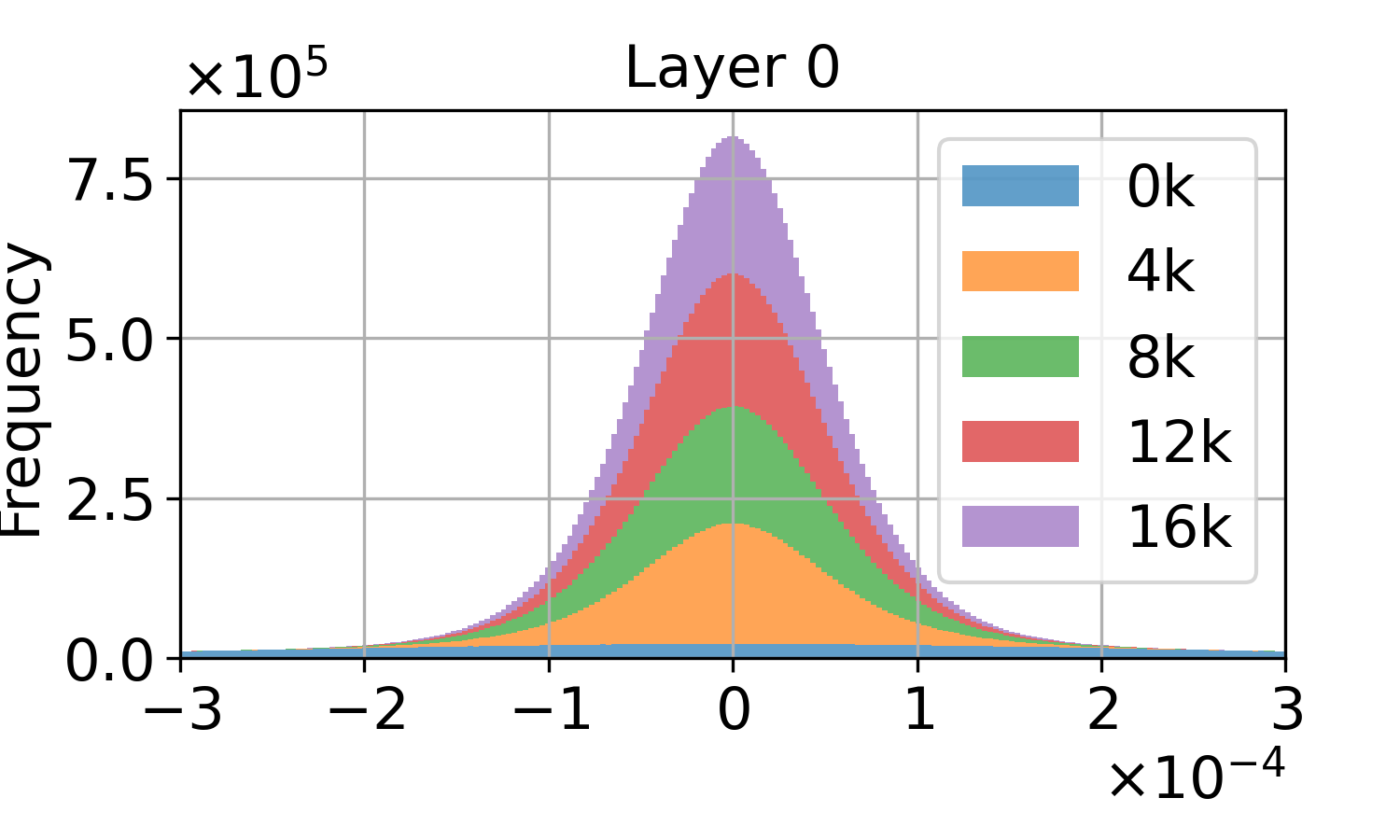}\hfill
    \includegraphics[width=0.48\linewidth]{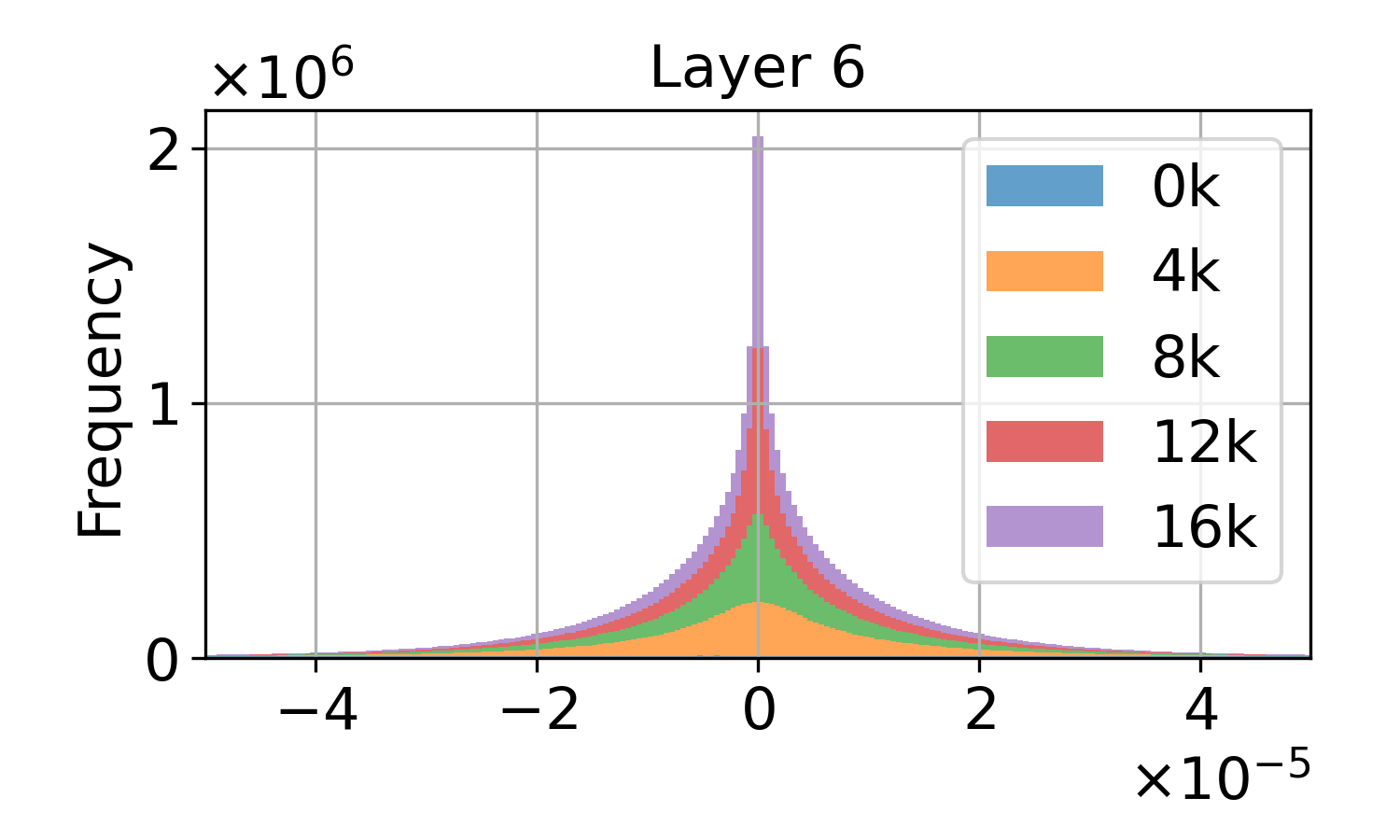}\hfill
    \includegraphics[width=0.48\linewidth]{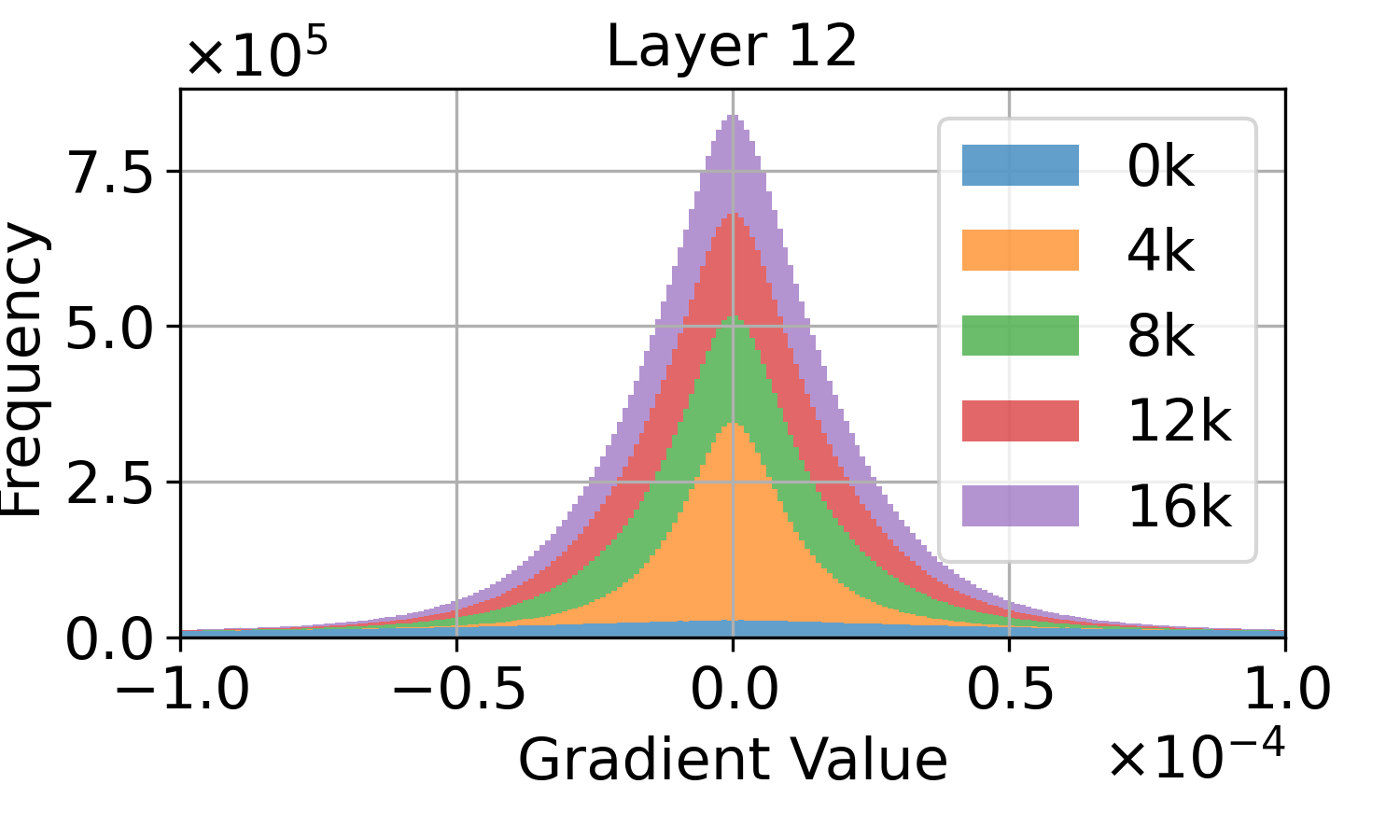}\hfill
    \includegraphics[width=0.48\linewidth]{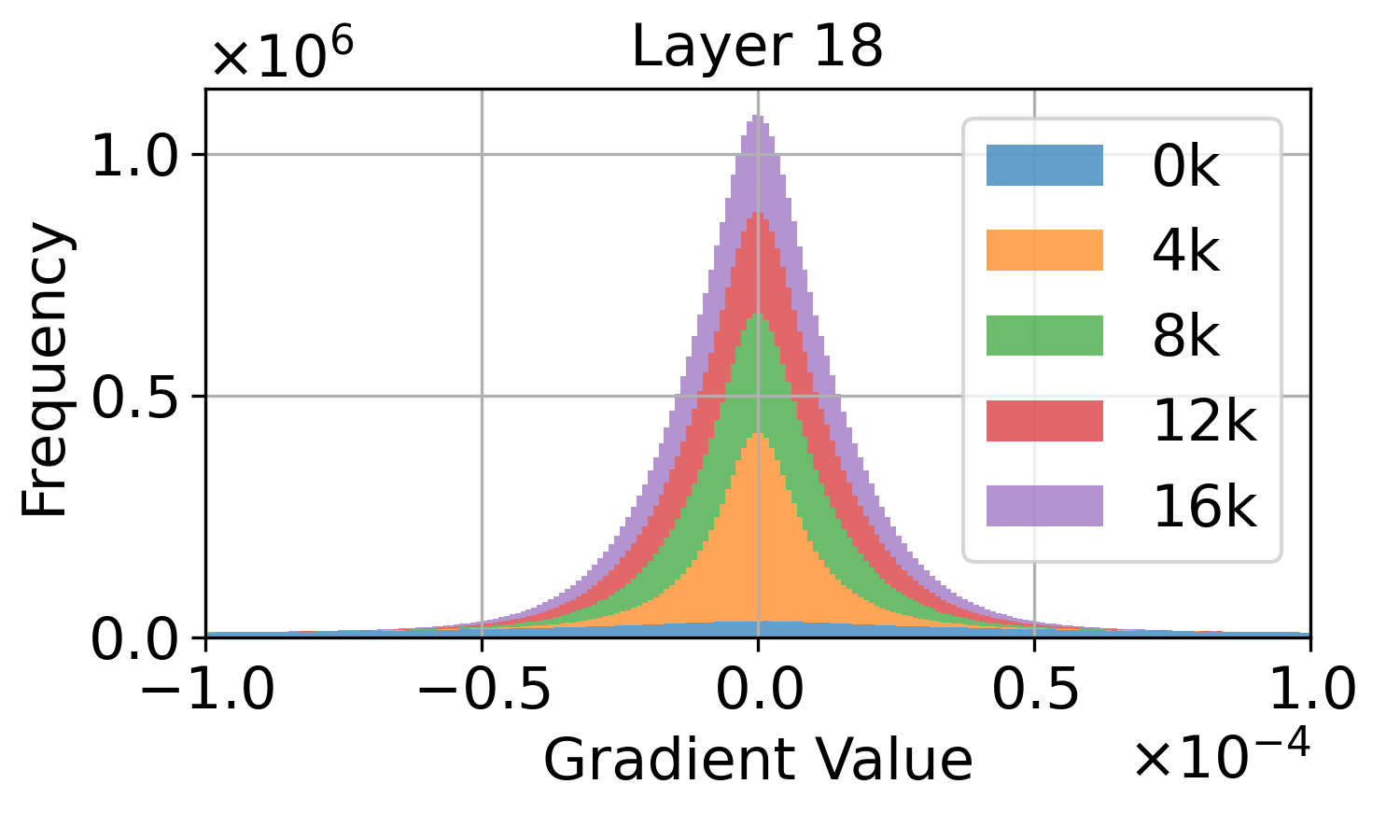}
    \caption{Gradient distribution of GPT2-345M across different model layers at different iterations.}
    \label{fig:layers_gradients.png}
    \vspace{-5mm}
\end{figure}

Figure \ref{fig:layers_gradients.png} tracks the gradient evolution of GPT2-345M across distinct layers (0--18) and intervals (0--16k iterations). Initially, the broad distribution reflects high uncertainty in feature extraction. As training advances, gradient values significantly narrow and exhibit zero-concentration. This reduction in variance indicates that the model is stabilizing and refining its learned features.
This convergence is driven by two factors. Primarily, as the model approaches local minima, the magnitude of gradients naturally diminishes. Secondarily, learning rate schedules, such as cosine annealing \cite{94}, progressively reduce the step size. This decay suppresses update volatility and further compacts the gradient distribution, creating favorable conditions for aggressive compression strategies (e.g., quantization or sparsification) in later training stages.

\emph{\textbf{Observation 3: Gradient matrices exhibit strong structural correlations early in training, which diminish over time.}}

\begin{figure}[tbp]
    \centering
    \subfloat[Random]{
        \includegraphics[width = 0.475\linewidth]{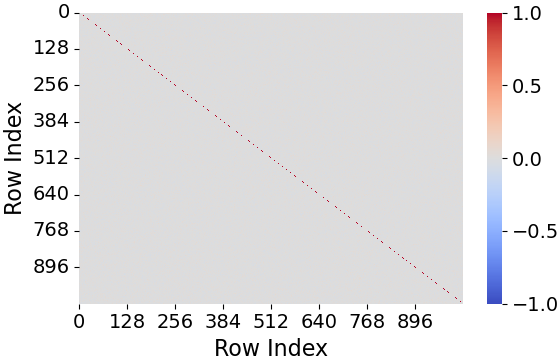}
        \label{fig:random_matrix}
    }\hfill
    \subfloat[Layer 0, 1k]{
        \includegraphics[width = 0.475\linewidth]{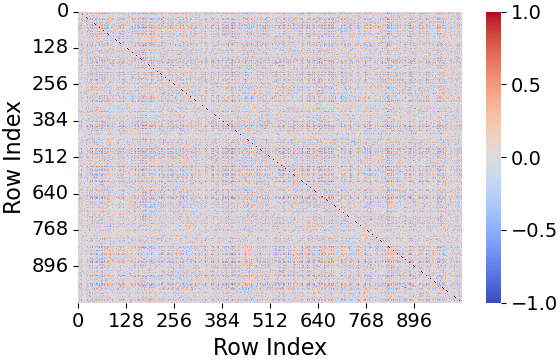}
        \label{fig:gradient_matrix_1}
    }
    
    \subfloat[Layer 11, 1k]{
        \includegraphics[width = 0.475\linewidth]{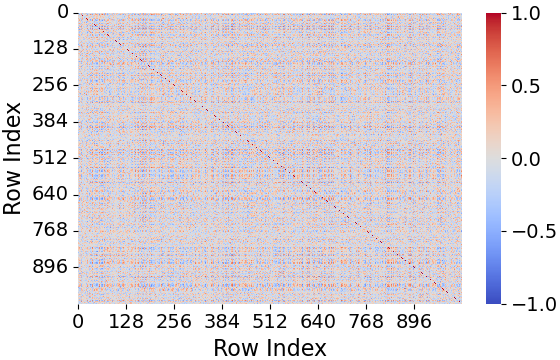}
        \label{fig:gradient_matrix_2}
    }\hfill
    \subfloat[Layer 0, 11k]{
        \includegraphics[width = 0.475\linewidth]{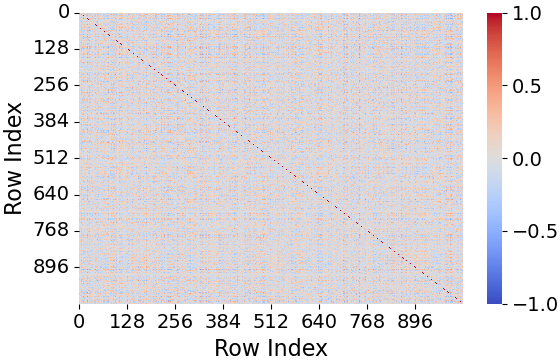}
        \label{fig:gradient_matrix_3}
    }
    \vspace{-2mm}
    \caption{Gradient matrix correlation of GPT2.}
    \vspace{-5mm}
    \label{fig:matrix_corr}
\end{figure}

Gradient correlations arise from the chain rule and parameter coupling. We visualized Pearson correlation heatmaps for GPT2-345M gradients. Early in training (1k iterations, Figure \ref{fig:gradient_matrix_1} \& \ref{fig:gradient_matrix_2}), layers exhibit strong correlations. This indicates that early updates are highly synergistic and structurally coupled, meaning that compression errors in one part could significantly propagate and disrupt the global training dynamics.
Conversely, as the model stabilizes (11k iterations, Figure \ref{fig:gradient_matrix_3}), correlations weaken. 
This decoupling suggests that gradient updates become more independent and localized. Consequently, the model becomes more robust to compression in later stages, further supporting the strategy of increasing compression rates as training progresses.

\subsection{Insights}
\label{sec:Motivation}

Observation 2 indicates concentration around zero rather than exact sparsity; directly pruning small values may therefore discard useful information~\cite{13}. Low-rank decomposition instead exploits the matrix structure observed in Observation 3 and preserves global information better than element-wise pruning. Meanwhile, a static rank~\cite{50} cannot follow the entropy evolution in Observation 1: it may be overly aggressive during the unstable phase and unnecessarily conservative after stabilization. These observations motivate EDGC, which uses real-time entropy to adjust compression ranks dynamically.

\section{Design of EDGC}
\label{sec: Design}

To fully exploit the evolving stability of LLM gradients, we propose EDGC, an entropy-driven dynamic gradient compression framework. Implementing dynamic compression poses three core challenges: (1) Efficiency: How to measure information content in massive gradients without slowing down training? (2) Theory: How to rigorously map information content (entropy) to an optimal compression rank? (3) System Integration: How to synchronize dynamic compression across complex 3D parallellism efficiently?

\begin{figure}[tbp]
     \centering
     \includegraphics[width=\linewidth]{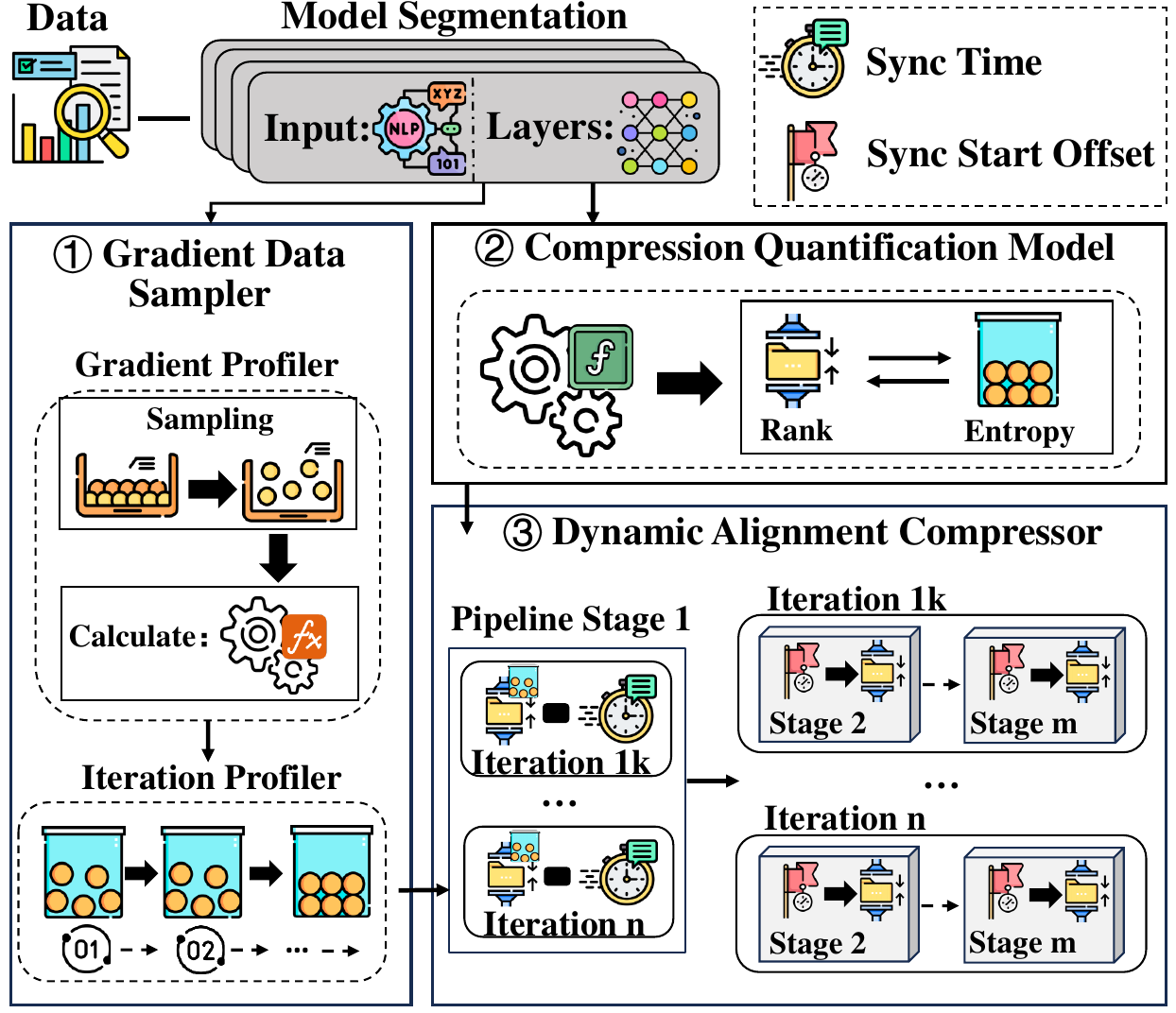}
     \vspace{-3mm}
     \caption{Overview of EDGC.}
     \vspace{-5mm}
     \label{fig:Overview}
\end{figure}

\begin{figure}[tbp]
     \centering
     \includegraphics[width=0.83\linewidth]{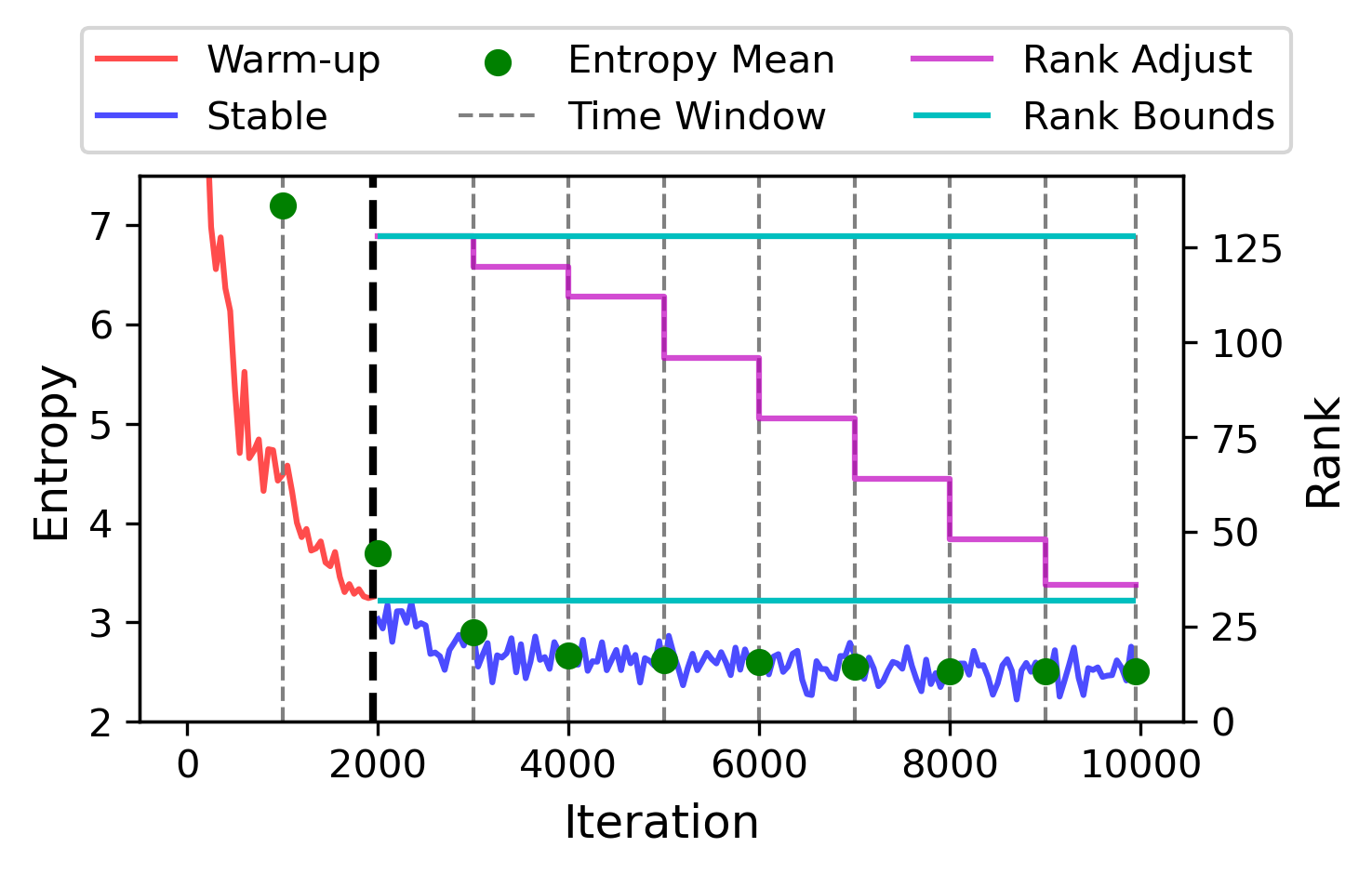}
     \vspace{-3mm}
     \caption{Visualization of warm-up, entropy calculations, rank bounds, and window-based compression adjustments. }
     \vspace{-3mm}
     \label{fig:overview1}
\end{figure}

As illustrated in Figure \ref{fig:Overview}, EDGC addresses these through three integrated components: a
gradient data sampler (GDS), a compression quantification
model (CQM), and a dynamic alignment compressor (DAC). Firstly, GDS employs a down-sampling technique that estimates gradient entropy from a subset of data, avoiding the high computational cost of full-scale calculation. Secondly, CQM formulates the relationship between gradient entropy and compression rank, enabling precise, data-driven rank adjustments. Finally, DAC manages the application of compression. Crucially, it incorporates a warm-up phase to bypass the high variability of initial gradients, and a window-based mechanism to adjust ranks periodically (rather than per-iteration), as illustrated in Figure \ref{fig:overview1}. This design minimizes computational overhead while ensuring consistent synchronization.

\subsection{Gradient Data Sampler (GDS)}
Considering the billions to trillions of parameters in LLMs, the main goal of GDS is to efficiently obtain a representative measure of gradients. It employs down-sampling at two levels: within a time window and during an iteration.

First, GDS evaluates gradients at a specified iteration sampling rate (ISR), denoted as $\alpha$. 
Specifically, within each time window, gradient variation is calculated once every $\frac{1}{\alpha}$ iterations.
Second, GDS selects a data subset during an iteration based on a specific gradient sampling rate (GSR), represented as $\beta$.
This approach significantly reduces the computational load, enabling rapid and accurate assessment of the gradient's condition while retaining essential information. As detailed in Appendix D of the Supplementary Document, our ablation studies show that GDS reduces entropy-estimation time by approximately 94\%.

\subsection{Compression Quantification Model (CQM)}

The precise quantification of the relationship between compression rank and gradient entropy remains underexplored, presenting a fundamental challenge that CQM aims to address.
Therefore, we introduce two established theorems to support the proof of our derived theorems. 
For space considerations, we defer all proofs of the lemmas and theorems to Appendix B of the Supplementary Document.

\textbf{Lemma 1. (CDF of the Marchenko-Pastur Distribution) }
Marcenko-Pastur Theorem \cite{96} describes the asymptotic behavior of eigenvalues of large random covariance matrices. Given a matrix $A\in \mathbb{R}^{m\times n}$ and the standard deviation of each entry in $A$ is $1$, the CDF of $AA^T$'s eigenvalue $\lambda$ satisfies:
\begin{align}
F(\lambda;m,n) = \frac{1}{2\pi m} F(\lambda; a, b)
\label{eq:CDF}
\end{align}
with $a = (\sqrt{n} - \sqrt{m})^2, b = (\sqrt{n} + \sqrt{m})^2$,
and
$$\begin{aligned}
    F(\lambda;a, b)&=-2\sqrt{ab} \arctan \left( \sqrt{\frac{b(\lambda - a)}{a(b - \lambda)}} \right) \\
    &+ (a + b) \arcsin \left( \sqrt{\frac{\lambda- a}{b - a}} \right) \\
    &+ \sqrt{(\lambda - a)(b - \lambda)}.
\end{aligned}$$

Based on this spectral distribution, we quantify the compression loss from low-rank approximation.

\label{sec:MCER}

\textbf{Theorem 1. (Relationship between Compression Rank and Compression Error.)}
Suppose that there is a random gradient matrix\footnote{LLM parameters are typically initialized to random values.} \(A \in \mathbb{R}^{m \times n}\) with entries that are i.i.d., having mean $0$ and variance $1$. When the compression rank is set to $r$, the squared compression error $||A-A_r||_F^2$ can be estimated as follows:
(a) Compute $a$ and $b$ from Lemma 1, and uniformly sample values $\{\lambda_0\}$ within the range from $a$ to $b$;
(b) Determine the corresponding cumulative probabilities $\{p_0\}$ based on the CDF $F(\lambda_0)$ in Lemma 1 to create pairs $\{(\lambda_0, p_0)\}$;
(c) Randomly sample $m$ values $\{p\}$ from a uniform distribution $(0, 1)$ to find the corresponding $\{\lambda\}$ from the pairs $\{(\lambda_0, p_0)\}$, where $\{\lambda\}$ represent the sampled eigenvalues;
(d) Calculate the sum of the smallest $m-r$ values in $\{\lambda\}$ as the estimation of the squared compression error.

This theorem provides a practical method for estimating the compression error $\epsilon = ||A-A_r||_F$ based on the compression rank $r$ and the dimensions $(m, n)$ of a random gradient matrix with standard normal entries. This process can be formalized as:
\begin{equation}
    \epsilon = g(r;m,n).
\label{eq:rank_vs_error}
\end{equation}

While our derivation assumes i.i.d. entries, this remains robust for real-world gradients for two reasons. First, spectral universality in high-dimensional probability ensures that the first two moments (mean and variance) dominate spectral properties, making the model a valid approximation even for non-Gaussian distributions. Second, real gradients exhibit structural correlations (Observation 3), causing singular values to decay faster than in i.i.d. matrices. Our model thus provides a conservative upper bound; if a rank $r$ satisfies the error threshold under the random assumption, the actual error on correlated gradients will be lower, providing a reliable safety margin throughout training.

\textbf{Constraint 1. (Constant absolute value of gradient matrix compression error.)} To facilitate the adjustment of compression ranks, we impose a fixed compression error $\epsilon_{ini}$ constraint during LLM training. This error is established when gradient compression is activated, as detailed in Section \ref{sec:warmup}.

\textbf{Theorem 2. (Relationship between Compression Rank and Standard Deviation.)} Let $\sigma_0$ denote the standard deviation of a random gradient matrix $A\in\mathbb{R}^{m\times n}$. The compression error $\epsilon=||A-A_r||_F$ adheres to Constraint 1 throughout the training process. If the standard deviation shifts from $\sigma_0$ to $\sigma_1$ during LLM training, the compression rank can be adjusted from $r_0$ to $r_1$ by:
   $$r_1 = g^{-1}\left(\frac{\sigma_0}{\sigma_1}g(r_0)\right).$$

\textbf{Lemma 2. (Relationship between Gradient Entropy and Standard Deviation.)}
For a random variable $X$ which follows a normal distribution with mean $\mu$ and standard deviation $\sigma$, its information entropy is given by:
\begin{equation}
    H(X) = \log \sigma + \frac{1}{2} \log 2\pi e.
\end{equation}

While actual gradient distributions may exhibit heavy-tailed characteristics, our reliance on the normal distribution in Lemma 2 is justified by the principle of maximum entropy. For a given variance, the normal distribution represents the "worst-case" uncertainty. This ensures our rank adaptation is conservative and robust, avoiding underestimation of information even in leptokurtic scenarios. As training stabilizes, decreasing kurtosis further aligns gradients with Gaussianity. Importantly, since EDGC tracks relative entropy trends rather than absolute values, this approximation guides compression dynamics across scales.

\textbf{Theorem 3. (Relationship between Compression Rank and Gradient Entropy)}
During the training process, suppose the gradient entropy of matrix $A$ changes from $H_0$ to $H_1$, the rank should be adjusted from $r_0$ to $r_1$ with a fixed error,  which satisfies:
\begin{equation}
    r_1 = g^{-1}\left(e^{H_0-H_1}g(r_0)\right)
\end{equation}

Based on this theoretical foundation, we propose a dynamic adjustment strategy. To mitigate the computational overhead of frequent rank updates (e.g., memory reallocation), we implement a window-based mechanism, balancing precise error control with system efficiency.

\subsection{Dynamic Alignment Compressor (DAC)}

As analyzed in Section \ref{sec:3D Parallelism}, two fundamental challenges in PP are identified: (1) temporal misalignment in parameter synchronization due to forward-backward dependencies, and (2) heterogeneous communication demands from uneven parameter distribution.
\paragraph{Computing Rank Bounds via Compression Rank and Communication Time.}
CQM defines the relationship between compression rank and DP communication performance by carefully selecting various ranks.
To achieve temporal benefits in compression and decompression, the total communication time $T_{com}$ for any rank must satisfy:
\begin{align}
    T_{com} = T_{compre} + \frac{D_{compre}}{B} + T_{decom} \leq \frac{D_{origi}}{B}
    \label{eq:diff}
\end{align}
where $T_{compre}$ denotes the time needed to compress data, while $D_{compre}$ indicates the amount of data after compression. $B$ represents the network bandwidth. Furthermore, $T_{decom}$ is the time required to decompress data, and $D_{origi}$ signifies the volume of the original data before compression.

\begin{figure}[tbp]
     \centering
     \includegraphics[width=0.88\linewidth]{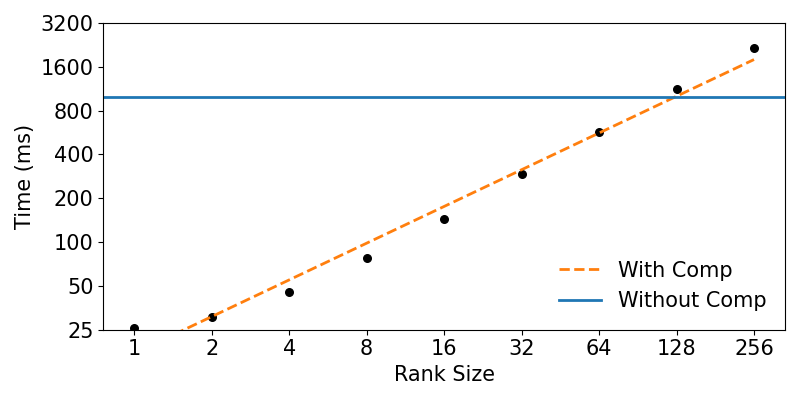}
     \vspace{-3mm}
     \caption{Communication time vs. rank values.}
     \label{fig:rank and time}
     \vspace{-5mm}
\end{figure}

If the inequality is violated, compression fails to reduce communication time, defining the upper bound $r_{max}$. To preserve accuracy, we also set a lower bound $r_{min}$, typically within $[\frac{r_{max}}{6}, \frac{r_{max}}{4}]$. Since hardware and network variations make static estimates imprecise, we use real-time data to ensure reliability.
Empirical data from pre-training GPT2-2.5B (Figure \ref{fig:rank and time}) in a 32 Gbps environment with TP=4, PP=4, and DP=2 reveals an approximately linear relationship between rank $r$ and communication time $T_{\mathrm{com}}(r)$. We therefore model it as:
\begin{align}
    T_{\mathrm{com}}(r)=\eta r
    \label{eq:5}
\end{align}
where $\eta$ reflects the influence of compression rank on communication time. 
Empirical evaluation within the rank bounds shows a mean absolute percentage error (MAPE) of only 2.85\%, demonstrating its high predictive accuracy.

\paragraph{Adaptive Warm-up Phase Determination.}
\label{sec:warmup}
As discussed in Section~\ref{sec:observation}, LLMs are sensitive to the precision of gradient updates during initial training. Early implementation of gradient compression can cause irreversible declines in model accuracy. To address this, EDGC initiates a warm-up phase where gradient compression is not applied. Additionally, it improves this process by adaptively defining the warm-up phase based on the compression error.

Specifically, before training, DAC estimates the initial maximum rank \( r_{max} \) and its error threshold \( \epsilon_{ini} \) using Eq. (\ref{eq:5}). During training, DAC computes the gradient entropy for each time window. A decrease in entropy signifies stabilization of the gradient distribution. Based on this change, DAC determines the new compression rank \(r_{\mathrm{new}}\) for the current time window using Equation (19) in Appendix B of the Supplementary Document. If \( r_{new} < r_{max} \), the warm-up phase ends, indicating that the gradient distribution has stabilized and the model is prepared for effective low-rank approximation.
We also impose an empirical constraint to ensure model stability. A 10\% warm-up phase is commonly used in practice to maintain convergence \cite{103}. If the calculated warm-up phase is under 10\% of the total iterations, we extend it to 10\%. This approach prevents premature compression and better aligns with actual training conditions.

\paragraph{Window-based Rank Adjustment.}
\label{sec:PCRAS}
\begin{figure}[tbp]
    \centering
    \subfloat[Layer 0]{
    \includegraphics[width = 0.48\linewidth]{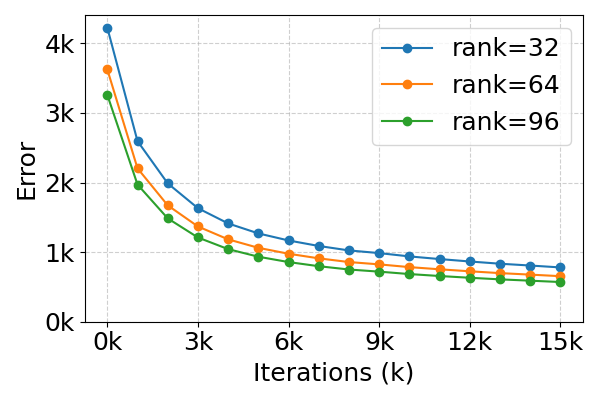}
    \label{fig:layer_0}
    }
    \subfloat[Layer 4]{
    \includegraphics[width = 0.48\linewidth]{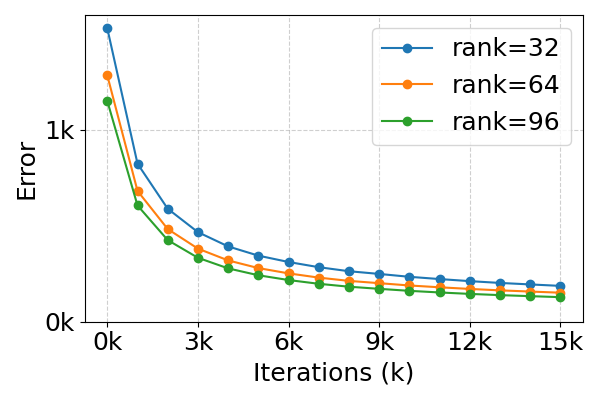}
    \label{fig:layer_1}
    }
\vspace{-2mm}
\caption{Changes in compression error vs. rank values.}
\label{fig:error_over_iteration}
\vspace{-5mm}
\end{figure}
Figure \ref{fig:error_over_iteration} presents our empirical analysis of gradient compression dynamics during GPT2-345M training. By monitoring layer-wise compression error, we observe three key phenomena. (1) Error decay property: the compression error under the same rank value gradually decreases. (2) Rank-error tradeoff: Under the same number of iterations, the smaller the rank, the larger the compression error. (3) Layer invariance: While absolute compression error magnitudes differ across layers, their relative error trends remain consistent.

DAC exploits these properties through a three-phase adaptation. Firstly, the compression error of the predefined maximum rank $r_{max}$ is calculated based on the initial gradients. Secondly, the average gradient entropy is computed in each time window $w$, and the compression direction is determined based on changes in the average entropy. If the average entropy decreases, indicating a more concentrated gradient distribution, the compression rank is reduced; otherwise, it is increased. Algorithm 1 in Appendix A of the Supplementary Document formalizes this adaptive process through entropy-driven rank adjustments constrained by empirically validated safety margins.

\paragraph{Stage-aligned Compression Rank Adjustment.}
\label{sec:SDC}
Building on the estimated communication time of Stage 1, the compression ranks of subsequent pipeline stages can be adjusted to align their communication completion times with that of Stage 1. Let the communication time of Stage 1 be denoted as $T_{\mathrm{com}}(r^{s_1}_{new})$. The compression ranks for other stages $i$ are then configured to maintain balance in the overall pipeline communication:
\begin{align}
    r^{s_i}_{new}=\frac{T_\mathrm{com}}\eta=\frac{T_{\mathrm{com}}(r^{s_1}_{new})+(i-1)*\overline{T}_{\mathrm{microBack}}}\eta
    \label{eq:16}
\end{align}
where $i$-1 represents the position offset between the $i$-th stage and Stage 1 in the pipeline. $\overline{T}_{\mathrm{microBack}}$ denotes the average time required for each stage to perform a micro-batch backpropagation.

To mitigate misalignment, it is essential to quantify and address time discrepancies between pipeline stages. The communication bottleneck occurs because Stage 1 initiates DP communication later than other stages due to sequential dependencies in backpropagation. By predicting the DP communication time for Stage 1 and the time differences at the start of communication for each stage, we can pinpoint the delay causing the bottleneck. Quantifying this misalignment enables DAC to adjust gradient compression ranks, balancing communication time across all stages. As illustrated in Algorithm 2 in Appendix A of the Supplementary Document, this strategy avoids an exclusive focus on Stage 1, which could introduce new bottlenecks, while preserving model accuracy by appropriately relaxing the compression ranks of subsequent stages, thereby improving overall pipeline efficiency.

\section{Evaluation}
\label{sec:evaluation}

We evaluate EDGC on GPT2 2.5B and 12.1B using 8- and 16-node clusters, respectively. Appendix C, Table 1 of the Supplementary Material summarizes the model, hardware, and network configurations.

\textbf{Baselines}. We compare EDGC against three state-of-the-art baselines: (1) Megatron-LM \cite{45} (no compression); (2) PowerSGD \cite{52} (low-rank decomposition); and (3) Optimus-CC \cite{35} (selection-phase compression with low-rank decomposition and error feedback).
To evaluate training efficiency and convergence, all methods undergo pre-training for over 230,000 iterations. 

\textbf{Model.} We chose GPT2 as it is a well-established benchmark and representative medium-scale LLM, commonly used in prior work (e.g., Optimus-CC) for fair comparison. Mid-scale LLMs are also widely adopted in real-world applications, making our evaluation relevant.

\textbf{Datasets.} We use the OpenWebText \cite{72} and OpenWebText2 \cite{39} datasets for pre-training. 
To evaluate the LLMs' generalization during training, we set aside 5\% of the dataset as a validation set.

\textbf{Metrics.} We use communication time and training time to evaluate the training speed. In addition, we assess the performance of LLMs after pre-training using perplexity (PPL) and loss values. PPL is one of the most common indicators for evaluating language models. These metrics are consistent with those used in mainstream models \cite{3}.

\subsection{Training Performance}

\paragraph{LLM Convergence.}

\begin{figure}[tbp]
    \centering
    \captionsetup[subfloat]{font=scriptsize}
    \subfloat[Loss change on GPT2-2.5B]{
    \includegraphics[width=0.48\linewidth]{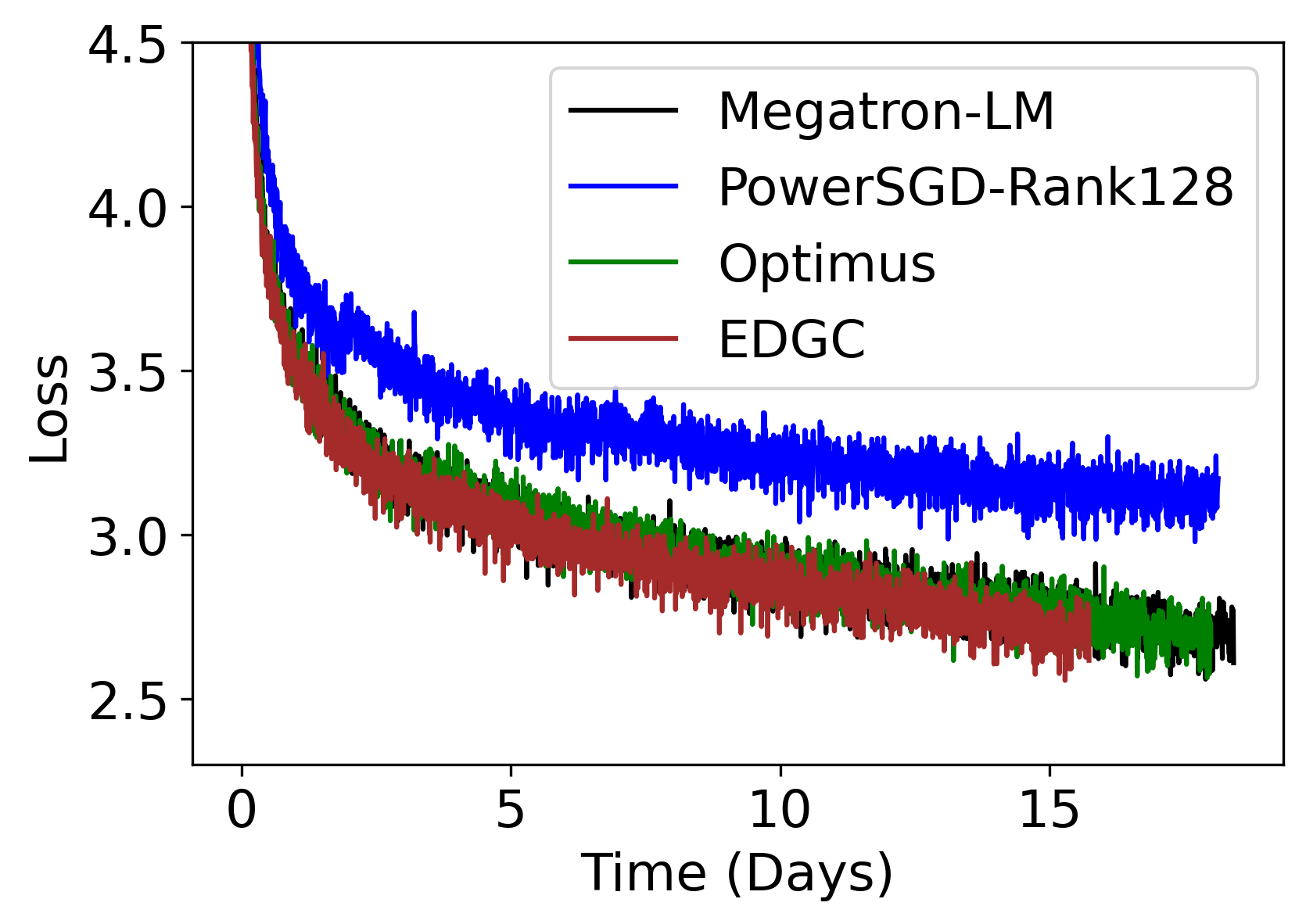}
    \label{fig:2.5B_loss_over_time}
    }
    \subfloat[Loss change on GPT2-12.1B]{
    \includegraphics[width=0.48\linewidth]{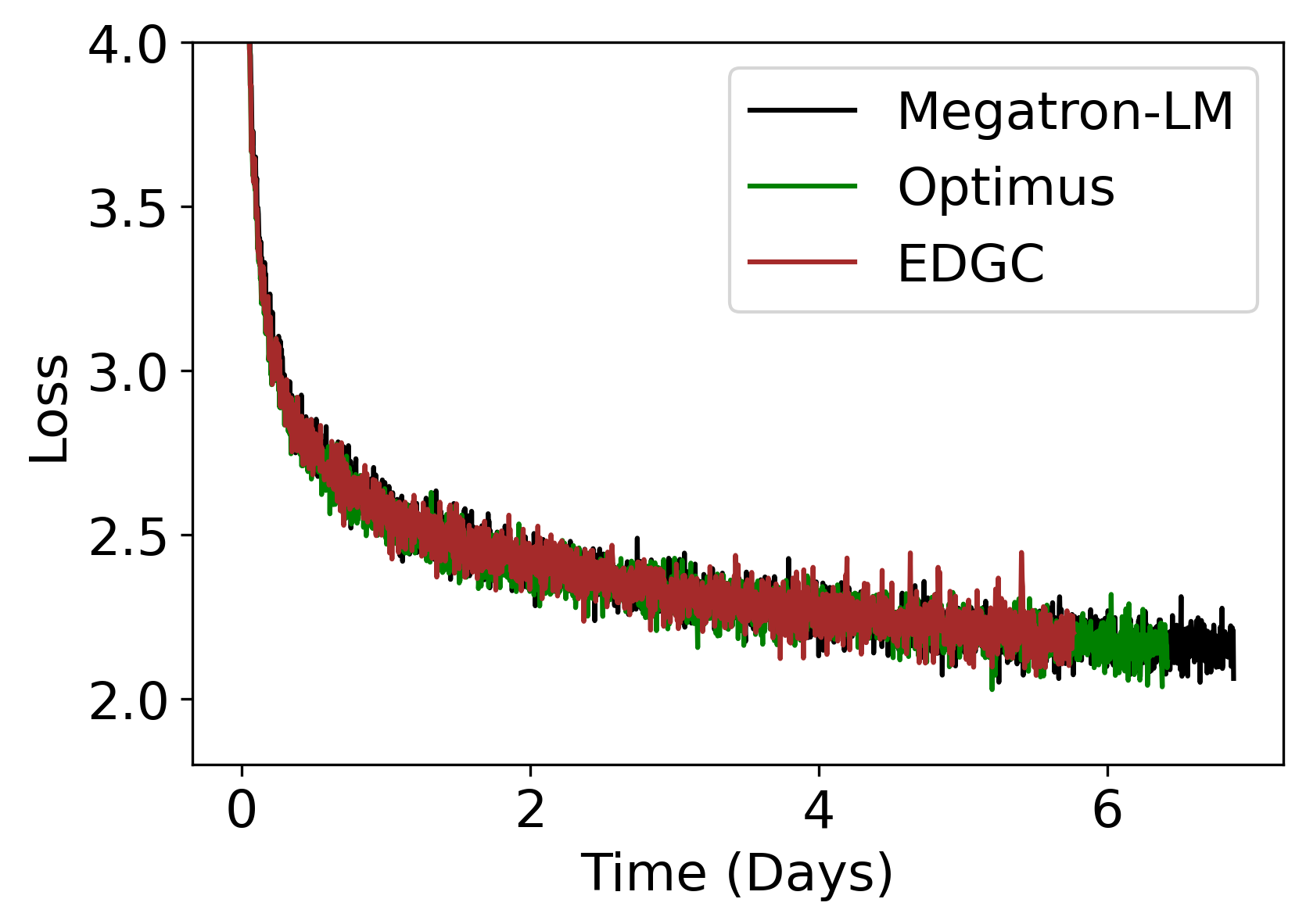}
    \label{fig:12.1B_loss_over_time}
    }
\vspace{-2mm}
\caption{The change in loss over time.}
\label{fig:loss_over_training}
\vspace{-2mm}
\end{figure}

Figure \ref{fig:2.5B_loss_over_time} and \ref{fig:12.1B_loss_over_time} illustrate the loss changes over time for GPT2-2.5B and GPT2-12.1B, respectively, under different methods. The results show that EDGC significantly outperforms the uncompressed scheme of Megatron-LM in terms of loss convergence speed, while achieving almost the same final loss value. These findings suggest that EDGC not only accelerates the convergence of LLMs, but also maintains strong generalization capability. Notably, in the GPT2-2.5B experiment, PowerSGD exhibited significantly worse performance, with both communication time and loss deviating considerably from those of Optimus-CC. 

\paragraph{Training Speedup.}

\begin{table}[t]
\centering
\caption{Training time and PPL after 230K iterations}
\vspace{-3mm}
\resizebox{\columnwidth}{!}{
\begin{tabular}{c|c|cccc}
\toprule
\textbf{Model} & \textbf{Metric} & \textbf{Megatron-LM} & \textbf{PowerSGD} & \textbf{Optimus-CC} & \textbf{EDGC} \\ \midrule
\multirow{2}{*}{\textbf{GPT2-2.5B}} 
& Time (day) & 18.44 & 18.14 & 18.01 & 15.74 \\
& PPL                 & 17.95 & 22.37 & 17.97 & 17.95 \\
\midrule
\multirow{2}{*}{\textbf{GPT2-12.1B}}
& Time (day) & 6.88  & -     & 6.42  & 5.77 \\
& PPL                 & 8.73  & -     & 8.84  & 8.87 \\
\bottomrule
\end{tabular}
}
\label{tab:Pretrain}
\vspace{-2mm}
\end{table}

Table \ref{tab:Pretrain} demonstrates that EDGC significantly accelerates training. For GPT2-2.5B, EDGC reduces training time by 12.60\% - 14.64\% across all baselines, cutting communication time by 45.8\% compared to Megatron-LM.  For GPT2-12.1B, EDGC achieves a 10.12\% - 16.13\% speedup and a 46.45\% reduction in communication. Notably, EDGC delivers consistent acceleration on both low-bandwidth (32 Gbps for GPT2-2.5B) and high-bandwidth (400 Gbps for GPT2-12.1B) networks.
Scalability is further verified on Llama-34B (32 GPUs, 400 Gbps). Even with conservative compression during early-stage training (10K iterations), EDGC reduces end-to-end time by 6\% and communication by 32.76\%, suggesting higher efficiency gains as training stabilizes.
Regarding model quality, EDGC matches Megatron-LM's PPL for GPT2-2.5B (17.95) and shows negligible increase for GPT2-12.1B ($\approx$0.1), unlike PowerSGD, which degrades to 22.37. This confirms that EDGC preserves model performance by leveraging gradient correlations (Observation 3). Even when trained to match baseline PPL, EDGC requires only 0.4\% - 2.1\% additional iterations, maintaining a net speedup of $\approx$14.3\%.

\begin{table}[t]
\centering
\caption{Accuracies on zero-shot tasks}
\vspace{-3mm}
\resizebox{\columnwidth}{!}{
\begin{tabular}{lcccccccc}
\toprule
\multirow{2}{*}{\textbf{Tasks}}  & \multicolumn{4}{c}{\textbf{GPT2-2.5B}} & \multicolumn{3}{c}{\textbf{GPT2-12.1B}} \\
\cmidrule(lr){2-5}\cmidrule(lr){6-8}
& \textbf{Megatron} & \textbf{Pow-SGD} & \textbf{Opti-CC} &  \textbf{EDGC} & \textbf{Mega} & \textbf{Opti-CC} & \textbf{EDGC} \\ \midrule
ARC\_easy~ & 41.75\% & 41.67\% & 41.73\% & 40.95\% & 60.65\% & 59.84\% & 58.25\% \\
ARC\_challenge~ & 21.84\% & 21.08\% & 21.84\% & 21.76\% & 28.84\% & 26.59\% & 27.47\% \\
HellaSwag~ & 35.55\% & 30.24\% & 35.41\% & 35.64\% & 42.34\% & 42.01\% & 40.65\% \\ 
OpenBookQA~ & 20.00\% & 17.40\% & 20.80\% & 21.20\% & 24.60\% & 25.01\% & 23.40\% \\
PIQA~ & 65.23\% & 63.28\% & 65.23\% & 65.23\% & 72.25\% & 70.50\% & 70.35\% \\ 
WinoGrande~ & 51.62\% & 50.99\% & 51.99\% & 52.25\% & 58.88\% & 57.89\% & 57.62\% \\\hline
\end{tabular}
}
\label{tab:zero_shot}
\vspace{-5mm}
\end{table}

\paragraph{Accuracies on Zero-shot Tasks.}

We evaluated generalizability across standard zero-shot benchmarks, including ARC, HellaSwag, OpenBookQA, PIQA, and WinoGrande (Table \ref{tab:zero_shot}). For GPT2-2.5B, EDGC matches Megatron-LM within training randomness. For GPT2-12.1B, performance is comparable, with minor fluctuations attributable to the limited training duration (230K iterations) adopted to align with Optimus-CC, rather than the full 500K required for full convergence. As training stabilizes and gradient entropy decreases, EDGC is expected to sustain high compression rates with minimal accuracy loss.

\subsection{Scope and Generalization}
\label{sec:discussion}

We examine EDGC across quality targets, model architectures, network conditions, and accelerator platforms. (1) To reach the PPL attained by the uncompressed baseline after 230K iterations, EDGC requires only 0.4\% and 2.1\% additional iterations for GPT2-2.5B and GPT2-12.1B, respectively. Communication savings nevertheless reduce the corresponding end-to-end training time by 14.30\% and 14.37\%. (2) EDGC also applies to large-scale MoE training. Although MoE routing introduces All-to-All communication, synchronizing expert parameters can still incur substantial All-Reduce overhead. Our Qwen3-32B experiments show that compressing this synchronization reduces training time. (3) The benefit increases under constrained bandwidth: when bandwidth decreases from 400 to 32~Gbps, communication grows from 34\% to 87\% of total training time, providing greater scope for compression. (4) Preliminary LLaMA-13B experiments on Hygon DCUs show approximately 20\% higher throughput at comparable convergence, consistent with the results on NVIDIA GPUs. A full long-run evaluation remains ongoing.

\section{Related Work}

Communication optimization for distributed LLM training broadly follows three directions. First, overlap methods such as WFBP~\cite{84} hide communication behind backpropagation. Their benefits diminish when communication dominates computation, and they depend on careful layer scheduling~\cite{32}. Second, semi-synchronous methods such as SSP~\cite{58} and SpecSync~\cite{91} reduce synchronization waits by allowing workers to proceed with stale gradients. However, staleness may degrade convergence and model quality, requiring careful tuning. Third, gradient compression directly reduces transferred data through quantization, including 1-bit Adam~\cite{87} and low-precision training~\cite{49}, or sparsification~\cite{28}. Large-scale systems such as DeepSeek-V3~\cite{95} further combine low-precision training with computation--communication overlap. Nevertheless, existing gradient-compression methods generally use static configurations that may be too aggressive early in training and unnecessarily conservative after gradients stabilize. EDGC instead adapts the compression rank to gradient evolution.

\section{Conclusions}
\label{sec:conclusion}
In this work, we propose EDGC, an entropy-driven dynamic gradient compression strategy aimed at enhancing communication efficiency in large, distributed models. Current communication compression methods often fail to adapt to changing training conditions, resulting in inefficient communication and possible declines in model quality. To tackle these challenges, we introduced several innovative techniques, including adaptive compression threshold estimation, a dynamic warm-up mechanism, periodic compression ratio adjustments, and a layered compression strategy. These techniques effectively reduce communication overhead during training while preserving model accuracy. More broadly, EDGC offers the community a practical step toward more resource-efficient LLM training by making communication compression adaptive to the evolving dynamics of optimization.

\section*{Limitations}

\paragraph{Evaluation coverage.}
Large-scale LLM pre-training is expensive, so our evaluation cannot exhaustively cover all model families, parameter scales, cluster sizes, and parallel configurations. The main experiments focus on GPT-2, while the LLaMA- and Qwen-based experiments provide additional evidence across model and hardware settings. Broader long-run evaluation on newer architectures, larger MoE configurations, and multilingual pre-training remains future work.

\paragraph{System dependency.}
EDGC provides the greatest benefit when gradient communication constitutes a substantial fraction of training time. Its speedup therefore depends on the hardware platform, network bandwidth and topology, parallel configuration, and the extent to which communication can already be overlapped with computation. The benefit may be smaller in compute-bound or extremely high-bandwidth environments.

\paragraph{Modeling assumptions.}
The entropy-to-rank model relies on random-matrix and maximum-entropy approximations. Although these assumptions provide a conservative basis for rank adjustment, they may not fully capture heavy-tailed or highly structured gradient distributions. EDGC also requires initial profiling and introduces nonzero overhead for gradient sampling, entropy estimation, and rank adjustment.

\paragraph{Adaptive robustness.}
EDGC adjusts compression ranks using sampled gradient entropy and window-level statistics. More complex training regimes, including long-context pre-training, instruction tuning, and reinforcement learning from human feedback, may exhibit different gradient dynamics and require different sampling rates, window sizes, rank bounds, or adaptation policies. Evaluating and automatically configuring EDGC for these regimes remains an important direction for future work.


\bibliography{custom}

@misc{103,
author={PyTorch},
title={{A}ccelerating {P}y{T}orch {D}{D}{P} by 10{X} {W}ith {P}ower{S}{G}{D}},
url={https://medium.com/pytorch/accelerating-pytorch-ddp-by-10x-with-powersgd-585aef12881d},
year={2021},
note={[Accessed 09-04-2025]},
}

@article{96,
  title={Rate of convergence in probability to the Marchenko-Pastur law},
  author={Friedrich G{\"o}tze and Alexander N. Tikhomirov},
  journal={Bernoulli},
  year={2004},
  volume={10},
  pages={503-548},
  url={https://api.semanticscholar.org/CorpusID:119916628}
}

@article{95,
  title={Deepseek-v3 technical report},
  author={Liu, Aixin and Feng, Bei and Xue, Bing and Wang, Bingxuan and Wu, Bochao and Lu, Chengda and Zhao, Chenggang and Deng, Chengqi and Zhang, Chenyu and Ruan, Chong and others},
  journal={arXiv preprint arXiv:2412.19437},
  year={2024}
}

@inproceedings{94,
  author={Ilya Loshchilov and Frank Hutter},
  title={{SGDR:} Stochastic Gradient Descent with Warm Restarts},
  booktitle={5th International Conference on Learning Representations (ICLR)},
  publisher={OpenReview.net},
  year={2017},
  url={https://openreview.net/forum?id=Skq89Scxx},
  bibsource={dblp computer science bibliography, https://dblp.org}
}

@INPROCEEDINGS{91,
  author={Zhang, Chengliang and Tian, Huangshi and Wang, Wei and Yan, Feng},
  booktitle={2018 IEEE 38th International Conference on Distributed Computing Systems (ICDCS)}, 
  title={Stay Fresh: Speculative Synchronization for Fast Distributed Machine Learning}, 
  year={2018},
  volume={},
  number={},
  pages={99-109},
  doi={10.1109/ICDCS.2018.00020}}

@inproceedings{87,
  author={Hanlin Tang and Shaoduo Gan and Ammar Ahmad Awan and Samyam Rajbhandari and Conglong Li and Xiangru Lian and Ji Liu and Ce Zhang and Yuxiong He},
  title={1-bit Adam: Communication Efficient Large-Scale Training with Adam's Convergence Speed},
  booktitle={Proceedings of the 38th International Conference on Machine Learning (ICML)},
  volume={139},
  pages={10118--10129},
  year={2021},
  url={http://proceedings.mlr.press/v139/tang21a.html}
}

@misc{85,
author={Wikimedia Foundation},
title={Wikimedia Downloads},
year={n.d.},
url={https://dumps.wikimedia.org},
note={[Accessed 09-04-2025]},
}

@inproceedings{84,
  author={Hao Zhang and Zeyu Zheng and Shizhen Xu and Wei Dai and Qirong Ho and Xiaodan Liang and Zhiting Hu and Jinliang Wei and Pengtao Xie and Eric P. Xing},
  title={Poseidon: An Efficient Communication Architecture for Distributed Deep Learning on {GPU} Clusters},
  booktitle={Proceedings of the 2017 {USENIX} Annual Technical Conference},
  pages={181--193},
  year={2017},
  url={https://www.usenix.org/conference/atc17/technical-sessions/presentation/zhang},
  bibsource={dblp computer science bibliography, https://dblp.org}
}

@inproceedings{83,
  author={Rasley, Jeff and Rajbhandari, Samyam and Ruwase, Olatunji and He, Yuxiong},
  title={DeepSpeed: System Optimizations Enable Training Deep Learning Models with Over 100 Billion Parameters},
  year={2020},
  url={https://doi.org/10.1145/3394486.3406703},
  doi={10.1145/3394486.3406703},
  booktitle={Proceedings of the 26th ACM SIGKDD International Conference on Knowledge Discovery \& Data Mining},
  pages={3505--3506},
  numpages={2}
}

@misc{72,
    title={OpenWebText Corpus},
    author={Gokaslan, Aaron and Cohen, Vanya and Pavlick, Ellie and Tellex, Stefanie},
    howpublished={\url{http://Skylion007.github.io/OpenWebTextCorpus}},
    year={2019},
    note = {[Accessed 09-04-2025]},
}

@InProceedings{70,
  title={Characterizing Implicit Bias in Terms of Optimization Geometry},
  author={Gunasekar, Suriya and Lee, Jason and Soudry, Daniel and Srebro, Nathan},
  booktitle={Proceedings of the 35th International Conference on Machine Learning},
  pages={1832--1841},
  year={2018},
  volume={80},
  series={Proceedings of Machine Learning Research},
  url={https://proceedings.mlr.press/v80/gunasekar18a.html}
}

@inproceedings{58,
author = {Ho, Qirong and Cipar, James and Cui, Henggang and Kim, Jin Kyu and Lee, Seunghak and Gibbons, Phillip B. and Gibson, Garth A. and Ganger, Gregory R. and Xing, Eric P.},
title = {More effective distributed ML via a Stale Synchronous Parallel parameter server},
year = {2013},
booktitle = {Proceedings of the 27th International Conference on Neural Information Processing Systems - Volume 1},
pages = {1223--1231},
numpages = {9},
series = {NIPS'13}
}

@inproceedings{52,
  author={Thijs Vogels and Sai Praneeth Karimireddy and Martin Jaggi},
  title={PowerSGD: Practical Low-Rank Gradient Compression for Distributed Optimization},
  booktitle={Advances in Neural Information Processing Systems},
  pages={14236--14245},
  year={2019},
  url={https://proceedings.neurips.cc/paper/2019/hash/d9fbed9da256e344c1fa46bb46c34c5f-Abstract.html},
  bibsource={dblp computer science bibliography, https://dblp.org}
}

@inproceedings{50,
  author={Frank Seide and Hao Fu and Jasha Droppo and Gang Li and Dong Yu},
  title={1-bit stochastic gradient descent and its application to data-parallel distributed training of speech DNNs},
  booktitle={15th Annual Conference of the International Speech Communication Association},
  pages={1058--1062},
  year={2014},
  url={https://doi.org/10.21437/Interspeech.2014-274},
  doi={10.21437/INTERSPEECH.2014-274},
  bibsource={dblp computer science bibliography, https://dblp.org}
}

@article{49,
  title={8-Bit Approximations for Parallelism in Deep Learning},
  author={Tim Dettmers},
  journal={CoRR},
  year={2015},
  volume={abs/1511.04561},
  url={https://api.semanticscholar.org/CorpusID:15201887}
}

@article{45,
  title={Megatron-LM: Training Multi-Billion Parameter Language Models Using Model Parallelism},
  author={Mohammad Shoeybi and Mostofa Patwary and Raul Puri and Patrick LeGresley and Jared Casper and Bryan Catanzaro},
  journal={ArXiv},
  year={2019},
  volume={abs/1909.08053},
  url={https://api.semanticscholar.org/CorpusID:202660670}
}

@article{43,
  title={Carbon Emissions and Large Neural Network Training},
  author={David A. Patterson and Joseph Gonzalez and Quoc V. Le and Chen Liang and Llu{\'i}s-Miquel Mungu{\'i}a and Daniel Rothchild and David R. So and Maud Texier and Jeff Dean},
  journal={ArXiv},
  year={2021},
  volume={abs/2104.10350},
  url={https://api.semanticscholar.org/CorpusID:233324338}
}

@inproceedings{42,
    title="{BERT}: Pre-training of Deep Bidirectional Transformers for Language Understanding",
    author="Devlin, Jacob  and
      Chang, Ming-Wei  and
      Lee, Kenton  and
      Toutanova, Kristina",
    booktitle="Proceedings of the 2019 Conference of the North {A}merican Chapter of the Association for Computational Linguistics: Human Language Technologies, Volume 1 (Long and Short Papers)",
    year="2019",
    url="https://aclanthology.org/N19-1423/",
    doi="10.18653/v1/N19-1423",
    pages="4171--4186",
}

@article{39,
title={The {P}ile: An 800GB Dataset of Diverse Text for Language Modeling},
author={Gao, Leo and Biderman, Stella and Black, Sid and Golding, Laurence and Hoppe, Travis and Foster, Charles and Phang, Jason and He, Horace and Thite, Anish and Nabeshima, Noa and Presser, Shawn and Leahy, Connor},
journal={arXiv preprint arXiv:2101.00027},
year={2020}
}

@inproceedings{35,
author = {Song, Jaeyong and Yim, Jinkyu and Jung, Jaewon and Jang, Hongsun and Kim, Hyung-Jin and Kim, Youngsok and Lee, Jinho},
title = {Optimus-CC: Efficient Large NLP Model Training with 3D Parallelism Aware Communication Compression},
year = {2023},
isbn = {9781450399166},
url = {https://doi.org/10.1145/3575693.3575712},
doi = {10.1145/3575693.3575712},
booktitle = {Proceedings of the 28th ACM International Conference on Architectural Support for Programming Languages and Operating Systems, Volume 2},
pages = {560--573},
numpages = {14},
series = {ASPLOS 2023}
}

@article{34,
  title={Language models are unsupervised multitask learners},
  author={Radford, Alec and Wu, Jeffrey and Child, Rewon and Luan, David and Amodei, Dario and Sutskever, Ilya and others},
  journal={OpenAI blog},
  volume={1},
  number={8},
  pages={9},
  year={2019}
}

@ARTICLE{32,
  author={Shi, Shaohuai and Chu, Xiaowen and Li, Bo},
  journal={IEEE Transactions on Parallel and Distributed Systems}, 
  title={MG-WFBP: Merging Gradients Wisely for Efficient Communication in Distributed Deep Learning}, 
  year={2021},
  volume={32},
  number={8},
  pages={1903-1917},
  doi={10.1109/TPDS.2021.3052862}
}

@ARTICLE{29,
  author={Qu, Zhihao and Guo, Song and Wang, Haozhao and Ye, Baoliu and Wang, Yi and Zomaya, Albert Y. and Tang, Bin},
  journal={IEEE Transactions on Mobile Computing}, 
  title={Partial Synchronization to Accelerate Federated Learning Over Relay-Assisted Edge Networks}, 
  year={2022},
  volume={21},
  number={12},
  pages={4502-4516},
  doi={10.1109/TMC.2021.3083154}
}

@inproceedings{28,
author = {Li, Shigang and Hoefler, Torsten},
title = {Near-optimal sparse allreduce for distributed deep learning},
year = {2022},
isbn = {9781450392044},
url = {https://doi.org/10.1145/3503221.3508399},
doi = {10.1145/3503221.3508399},
booktitle = {Proceedings of the 27th ACM SIGPLAN Symposium on Principles and Practice of Parallel Programming},
pages = {135--149},
numpages = {15},
series = {PPoPP '22}
}

@inproceedings{27,
author = {Chen, Chia-Yu and Ni, Jiamin and Lu, Songtao and Cui, Xiaodong and Chen, Pin-Yu and Sun, Xiao and Wang, Naigang and Venkataramani, Swagath and Srinivasan, Vijayalakshmi and Zhang, Wei and Gopalakrishnan, Kailash},
title = {ScaleCom: scalable sparsified gradient compression for communication-efficient distributed training},
year = {2020},
isbn = {9781713829546},
booktitle = {Proceedings of the 34th International Conference on Neural Information Processing Systems},
pages={13551--13563},
series = {NIPS '20}
}

@misc{26,
author={Andrew Gibiansky},
title={{B}ringing {H}{P}{C} {T}echniques to {D}eep {L}earning},
howpublished={\url{https://andrew.gibiansky.com/blog/machine-learning/baidu-allreduce}},
year={2017},
note = {[Accessed 09-04-2025]},
}

@INPROCEEDINGS{16,
  author={Xu, Qifan and You, Yang},
  booktitle={2023 IEEE International Parallel and Distributed Processing Symposium (IPDPS)}, 
  title={An Efficient 2D Method for Training Super-Large Deep Learning Models}, 
  year={2023},
  volume={},
  number={},
  pages={222-232},
  doi={10.1109/IPDPS54959.2023.00031}
}

@inproceedings{13,
  author={Kang Zhao and Sida Huang and Pan Pan and Yinghan Li and Yingya Zhang and Zhenyu Gu and Yinghui Xu},
  title={Distribution Adaptive {INT8} Quantization for Training CNNs},
  booktitle={Thirty-Fifth {AAAI} Conference on Artificial Intelligence (AAAI)},
  pages={3483--3491},
  year={2021},
  url={https://doi.org/10.1609/aaai.v35i4.16462},
  doi={10.1609/AAAI.V35I4.16462}
}

@inproceedings{10,
author = {Li, Mu and Andersen, David G. and Smola, Alexander and Yu, Kai},
title = {Communication efficient distributed machine learning with the parameter server},
year = {2014},
booktitle = {Proceedings of the 28th International Conference on Neural Information Processing Systems - Volume 1},
pages = {19--27},
numpages = {9},
series = {NIPS'14}
}

@inproceedings{6,
author = {Dean, Jeffrey and Corrado, Greg S. and Monga, Rajat and Chen, Kai and Devin, Matthieu and Le, Quoc V. and Mao, Mark Z. and Ranzato, Marc'Aurelio and Senior, Andrew and Tucker, Paul and Yang, Ke and Ng, Andrew Y.},
title = {Large scale distributed deep networks},
year = {2012},
booktitle = {Proceedings of the 26th International Conference on Neural Information Processing Systems - Volume 1},
pages = {1223--1231},
numpages = {9},
series = {NIPS'12}
}

@article{4,
  author={Abhimanyu Dubey and Abhinav Jauhri and Abhinav Pandey and Abhishek Kadian and Ahmad Al{-}Dahle and Aiesha Letman and Akhil Mathur and Alan Schelten and Amy Yang and Angela Fan and others},
  title={The Llama 3 Herd of Models},
  journal={CoRR},
  volume={abs/2407.21783},
  year={2024},
  url={https://doi.org/10.48550/arXiv.2407.21783},
  doi={10.48550/ARXIV.2407.21783},
  eprinttype={arXiv},
  eprint={2407.21783}
}

@article{3,
  author={OpenAI},
  title={{GPT-4} Technical Report},
  journal={CoRR},
  volume={abs/2303.08774},
  year={2023},
  url={https://doi.org/10.48550/arXiv.2303.08774},
  doi={10.48550/ARXIV.2303.08774},
  eprinttype={arXiv},
  eprint={2303.08774}
}

@inproceedings{2,
  author={Yanping Huang and Youlong Cheng and Ankur Bapna and Orhan Firat and Dehao Chen and Mia Xu Chen and HyoukJoong Lee and Jiquan Ngiam and Quoc V. Le and Yonghui Wu and Zhifeng Chen},
  title={GPipe: Efficient Training of Giant Neural Networks using Pipeline Parallelism},
  booktitle={Advances in Neural Information Processing Systems 32: Annual Conference on Neural Information Processing Systems},
  pages={103--112},
  year={2019},
  url={https://proceedings.neurips.cc/paper/2019/hash/093f65e080a295f8076b1c5722a46aa2-Abstract.html},
  bibsource={dblp computer science bibliography, https://dblp.org}
}

@article{1,
  title={Communication-Efficient Distributed Deep Learning: A Comprehensive Survey},
  author={Zhenheng Tang and Shaohuai Shi and Xiaowen Chu and Wei Wang and Bo Li},
  journal={ArXiv},
  year={2020},
  volume={abs/2003.06307},
  url={https://api.semanticscholar.org/CorpusID:212718058}
}

\appendix

\section{Algorithms of EDGC}

\begin{algorithm}[h]
\caption{Window-Based Dynamic Rank Adjustment}
\label{alg:window-rank}
\KwIn{Previous-window rank $r_{\mathrm{prev}}$; rank bounds
$[r_{\min},r_{\max}]$; initial compression error $\epsilon_{\mathrm{ini}}$
at $r_{\max}$}
\KwOut{Updated rank $r_{\mathrm{new}}^{s_1}$ for pipeline stage 1 in window
$w$ and predicted communication time
$T_{\mathrm{com}}(r_{\mathrm{new}}^{s_1})$}
Configure the adjustment limit $s$\;
Calculate the target rank $r_{\mathrm{new}}$ from the current gradient
entropy using Equation~\ref{eq:rank-entropy}, subject to
$\epsilon_{\mathrm{ini}}$\;
\eIf{$|r_{\mathrm{new}}-r_{\mathrm{prev}}|>s$}{
  \eIf{$r_{\mathrm{new}}>r_{\mathrm{prev}}$}{
    $r_{\mathrm{new}}^{s_1}\leftarrow r_{\mathrm{prev}}+s$\;
  }{
    $r_{\mathrm{new}}^{s_1}\leftarrow r_{\mathrm{prev}}-s$\;
  }
}{
  $r_{\mathrm{new}}^{s_1}\leftarrow r_{\mathrm{new}}$\;
}
$r_{\mathrm{new}}^{s_1}\leftarrow
\max\{r_{\min},\min\{r_{\max},r_{\mathrm{new}}^{s_1}\}\}$\;
Predict $T_{\mathrm{com}}(r_{\mathrm{new}}^{s_1})$ using the
communication-time model defined in the main paper\;
\Return{$r_{\mathrm{new}}^{s_1}$ and
$T_{\mathrm{com}}(r_{\mathrm{new}}^{s_1})$}
\end{algorithm}

\begin{algorithm}[h]
\caption{Stage-Aligned Dynamic Rank Adjustment}
\label{alg:stage-rank}
\KwIn{$T_{\mathrm{com}}(r_{\mathrm{new}}^{s_1})$; mean micro-batch
backward time $\overline{T}_{\mathrm{microBack}}$; number of pipeline
stages $S$; rank bounds $[r_{\min},r_{\max}]$}
\KwOut{Compression rank $r_{\mathrm{new}}^{s_i}$ for every stage
$i\in\{1,\ldots,S\}$}
$r_{\mathrm{new}}^{s_1}\leftarrow$ the stage-1 rank returned by
Algorithm~\ref{alg:window-rank}\;
\For{$i\leftarrow 2$ \KwTo $S$}{
  $T_{\mathrm{com}}^{s_i}\leftarrow
  T_{\mathrm{com}}(r_{\mathrm{new}}^{s_1})
  +(i-1)\overline{T}_{\mathrm{microBack}}$\;
  Compute $r_{\mathrm{new}}^{s_i}$ from
  $T_{\mathrm{com}}^{s_i}$ using the communication-time model defined
  in the main paper\;
  $r_{\mathrm{new}}^{s_i}\leftarrow
  \max\{r_{\min},\min\{r_{\max},r_{\mathrm{new}}^{s_i}\}\}$\;
}
\Return{$\{r_{\mathrm{new}}^{s_i}\}_{i=1}^{S}$}
\end{algorithm}

\section{Detailed Derivations for CQM}
\label{sec:cqm-derivations}

\subsection{CDF of the Marchenko--Pastur Distribution}

\textbf{Lemma 1.}
Let $A\in\mathbb{R}^{m\times n}$ have independent entries with zero mean
and unit variance. Under the Marchenko--Pastur model, define
\begin{equation}
a=(\sqrt{n}-\sqrt{m})^2,\qquad
b=(\sqrt{n}+\sqrt{m})^2.
\end{equation}
For $\lambda\in[a,b]$, the cumulative term used by CQM is
\begin{equation}
F(\lambda;m,n)=\frac{1}{2\pi m}\mathcal{F}(\lambda;a,b),
\label{eq:mp-cdf}
\end{equation}
where
\begin{align}
\mathcal{F}(\lambda;a,b)
={}&-2\sqrt{ab}\arctan\left[
\frac{b(\lambda-a)}{a(b-\lambda)}\right]^{1/2}\nonumber\\
&+(a+b)\arcsin\left[
\frac{\lambda-a}{b-a}\right]^{1/2}\nonumber\\
&+\sqrt{(\lambda-a)(b-\lambda)}.
\label{eq:mp-cdf-expanded}
\end{align}

\begin{proof}
Consider the indefinite integral
\begin{equation}
I=\int\frac{\sqrt{(b-\lambda)(\lambda-a)}}{\lambda}\,d\lambda.
\end{equation}
Let $y=\lambda-a$ and $c=b-a$. Then
\begin{equation}
I=\int\frac{\sqrt{(c-y)y}}{y+a}\,dy.
\end{equation}
Using $y=c\sin^2(t)$ gives
$dy=c\sin(2t)\,dt$. After setting $c(d-1)=2a$, the integral becomes
\begin{equation}
I=c\int\frac{1-\cos^2(2t)}{d-\cos(2t)}\,dt.
\end{equation}
Decomposing the numerator yields
\begin{align}
I={}&c(1-d^2)\int\frac{dt}{d-\cos(2t)}
+cd\int dt+c\int\cos(2t)\,dt.
\label{eq:mp-intermediate}
\end{align}
For the first integral, let $u=\tan(t)$. Since
$\cos(2t)=(1-u^2)/(1+u^2)$ and $dt=du/(1+u^2)$,
\begin{align}
\int\frac{dt}{d-\cos(2t)}
&=\int\frac{du}{(d+1)u^2+d-1}\nonumber\\
&=\frac{1}{\sqrt{d^2-1}}
\arctan\left(\sqrt{\frac{d+1}{d-1}}\tan t\right).
\end{align}
Substitution into Equation~\ref{eq:mp-intermediate} gives
\begin{align}
I={}&-c\sqrt{d^2-1}
\arctan\left(\sqrt{\frac{(d+1)y}{(d-1)(c-y)}}\right)\nonumber\\
&+cd\arcsin\left(\sqrt{\frac{y}{c}}\right)
+\sqrt{y(c-y)}.
\end{align}
Finally, substituting $y=\lambda-a$, $c=b-a$, and
$c(d-1)=2a$ yields Equation~\ref{eq:mp-cdf-expanded}.
\end{proof}

\subsection{Compression Rank and Compression Error}

\textbf{Theorem 1.}
Suppose that $A\in\mathbb{R}^{m\times n}$ has independent entries with
zero mean and unit variance. When the compression rank is $r$, the
squared compression error $\lVert A-A_r\rVert_F^2$ can be estimated by:
\begin{enumerate}
\item computing $a$ and $b$ from Lemma 1 and sampling candidate
eigenvalues $\{\lambda_0\}$ over $[a,b]$;
\item evaluating their cumulative probabilities
$\{p_0=F(\lambda_0;m,n)\}$;
\item sampling $m$ probabilities from the uniform distribution on
$(0,1)$ and mapping them through the numerical inverse CDF; and
\item summing the smallest $m-r$ sampled eigenvalues.
\end{enumerate}

\begin{proof}
The squared singular values of $A$ equal the eigenvalues of $AA^\top$.
For the best rank-$r$ approximation $A_r$, the
Eckart--Young--Mirsky theorem gives
\begin{equation}
\lVert A-A_r\rVert_F^2=\sum_{i=r+1}^{m}\lambda_i.
\label{eq:rank-error-sum}
\end{equation}
The Marchenko--Pastur model in Equation~\ref{eq:mp-cdf} provides the
asymptotic eigenvalue distribution of $AA^\top$. Sampling through its
numerical inverse CDF approximates the eigenvalues in
Equation~\ref{eq:rank-error-sum}; consequently, the sum of the smallest
$m-r$ sampled values estimates the squared compression error.
\end{proof}

Define the resulting rank-to-error mapping as
\begin{equation}
\epsilon=g(r;m,n).
\label{eq:rank-error-map}
\end{equation}

\subsection{Compression Rank and Standard Deviation}

\textbf{Theorem 2.}
Let $\sigma_0$ denote the standard deviation of a random gradient matrix
$A\in\mathbb{R}^{m\times n}$. Under the fixed absolute compression-error
constraint used by EDGC, if the standard deviation changes from
$\sigma_0$ to $\sigma_1$, the compression rank changes from $r_0$ to
$r_1$ according to
\begin{equation}
r_1=g^{-1}\left(\frac{\sigma_0}{\sigma_1}g(r_0)\right).
\label{eq:rank-standard-deviation}
\end{equation}

\begin{proof}
For matrices whose scale changes from $\sigma_0$ to $\sigma_1$,
\begin{equation}
\frac{\lVert A'\rVert_F}{\lVert A\rVert_F}
=\frac{\sigma_1}{\sigma_0}.
\end{equation}
At a fixed rank, the compression error scales with the matrix standard
deviation:
\begin{equation}
\epsilon\propto g(r;m,n)\sigma.
\end{equation}
Maintaining a constant absolute compression error therefore requires
\begin{equation}
g(r_0)\sigma_0=g(r_1)\sigma_1.
\end{equation}
Solving for $r_1$ gives
Equation~\ref{eq:rank-standard-deviation}.
\end{proof}

\subsection{Gradient Entropy and Dynamic Rank}

For a continuous random variable $X$ with density $f$, its differential
entropy is
\begin{equation}
H(X)=-\int f(x)\log f(x)\,dx.
\label{eq:entropy-definition}
\end{equation}

\textbf{Lemma 2.}
If $X$ follows a normal distribution with mean $\mu$ and standard
deviation $\sigma$, then
\begin{equation}
H(X)=\log\sigma+\frac{1}{2}\log(2\pi e).
\label{eq:normal-entropy}
\end{equation}

\begin{proof}
Substituting the normal density into
Equation~\ref{eq:entropy-definition} gives
\begin{align}
H(X)
={}&\int f(x)\log(\sqrt{2\pi}\sigma)\,dx\nonumber\\
&+\int f(x)\frac{(x-\mu)^2}{2\sigma^2}\,dx\nonumber\\
={}&\log(\sqrt{2\pi}\sigma)
+\frac{1}{2}\int y^2\frac{1}{\sqrt{2\pi}}
e^{-y^2/2}\,dy,
\end{align}
where $y=(x-\mu)/\sigma$. The remaining integral is the second moment
of a standard normal random variable and equals one, which yields
Equation~\ref{eq:normal-entropy}.
\end{proof}

As observed in the main paper, gradient distributions become more
concentrated as training progresses and can be approximated by a normal
distribution. Under this assumption, Lemma 2 links entropy directly to
the standard deviation.

\textbf{Theorem 3.}
Suppose the gradient entropy changes from $H_0$ to $H_1$. Under a fixed
absolute compression error, the rank changes from $r_0$ to $r_1$ as
\begin{equation}
r_1=g^{-1}\left(e^{H_0-H_1}g(r_0)\right).
\label{eq:rank-entropy}
\end{equation}

\begin{proof}
By Lemma 2,
\begin{equation}
H_1-H_0=\log\frac{\sigma_1}{\sigma_0},
\end{equation}
and hence
\begin{equation}
\frac{\sigma_0}{\sigma_1}=e^{H_0-H_1}.
\end{equation}
Substitution into Equation~\ref{eq:rank-standard-deviation} yields
Equation~\ref{eq:rank-entropy}.
\end{proof}

\section{Experimental Configurations}
\label{sec:configurations}

\begin{table}
\footnotesize
\centering
\caption{Experimental environments}\label{tab:environment}
\resizebox{\columnwidth}{!}{
\begin{tabular}{cccc}
\toprule
\multirow{6}{*}{\textbf{Cluster 1}} & \multirow{4}{*}{\makecell{Node}} & Number & 8 \\
&& CPU & Xeon Platinum 8163, 32 cores \\
&& Memory & 256 GB \\
&& GPU & 4$\times$ Nvidia Tesla V100 (32 GB) \\

\cmidrule(lr){3-4}
&\multirow{2}{*}{Interconnect} & Intra-node & NVLink (300 Gbps / GPU)\\
&& Inter-node & Ethernet (32 Gbps) \\
\midrule

\multirow{6}{*}{\textbf{Cluster 2}} & \multirow{4}{*}{\makecell{Node}} & Number & 16 \\
&& CPU & Xeon Platinum 8558, 48 cores \\
&& Memory & 512 GB \\
&& GPU & 4$\times$ Nvidia H100 (80 GB) \\

\cmidrule(lr){3-4}
&\multirow{2}{*}{Interconnect} & Intra-node & NVLink (900 Gbps / GPU)\\
&& Inter-node & Infiniband NDR (400 Gbps) \\
\midrule

\multirow{10}{*}{\textbf{Models}} & \multirow{3}{*}{Common} & Micro-batch  & 4 \\
&& Total mini-batch & 64 / 128  \\
&& Iterations & 230,000 \\

\cmidrule(lr){3-4}
&\multirow{3}{*}{\makecell{GPT2-2.5B}} & Layers &  52 \\
&& Hidden dim & 1,920 \\
&& Ways & TP4 / DP2 / PP4 \\

\cmidrule(lr){3-4}
&\multirow{3}{*}{\makecell{GPT2-12.1B}} & Layers &  76 \\
&& Hidden dim & 3,584 \\
&& Ways & TP4 / DP4 / PP4 \\

\bottomrule
\end{tabular}
}
\end{table}

Table~\ref{tab:environment} summarizes the node, network, and model
configurations used in the observation and evaluation sections of the
main paper. EDGC consists of three modules: GDS, CQM, and DAC. The
following experiments isolate the effects of these modules.

\section{Additional Ablation Studies}

\subsection{Analysis of GDS}
\label{sec:ablation-gds}

\begin{figure}[tb]
    \centering
    \includegraphics[width=0.85\linewidth]{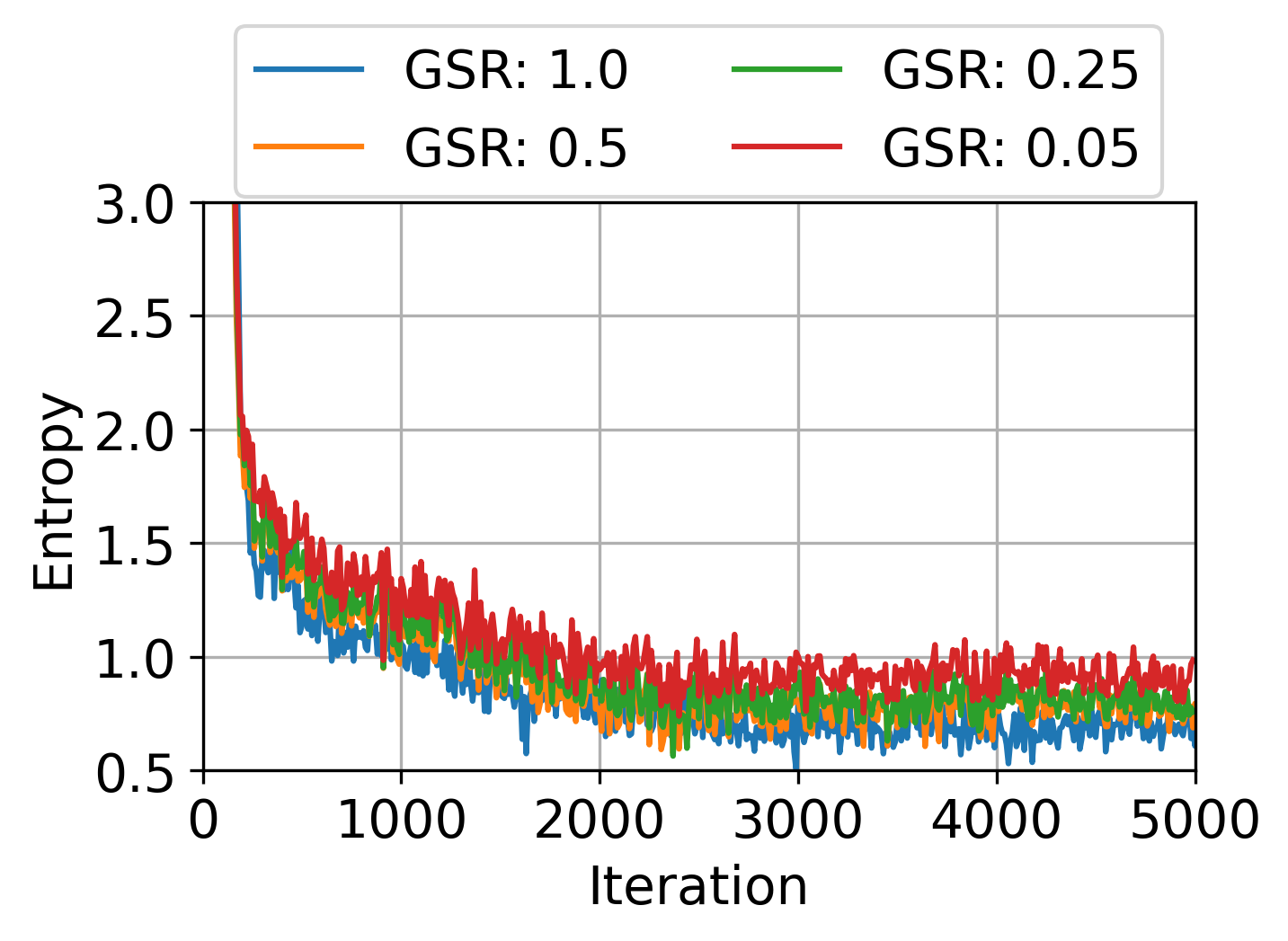}
    \vspace{-3mm}
    \caption{Gradient entropy under different gradient sampling rates.}
    \label{fig:gradient-sample}
    \vspace{-4mm}
\end{figure}

To evaluate the effect of sampling rates on entropy estimation and
computational efficiency, we experimented with gradient sampling rates
(GSR) $\beta\in\{0.05,0.25,0.5,1.0\}$ and iteration sampling rates
(ISR) $\alpha\in\{0.05,0.1,0.25,0.5\}$ using BERT-345M. We first
computed gradient entropy over 5,000 iterations using
Equation~\ref{eq:entropy-definition}. As shown in
Figure~\ref{fig:gradient-sample}, even small values of $\beta$ capture
the entropy dynamics, while $\beta=0.25$ provides a useful trade-off
between accuracy and overhead.

\begin{figure}[tb]
    \centering
    \includegraphics[width=0.85\linewidth]{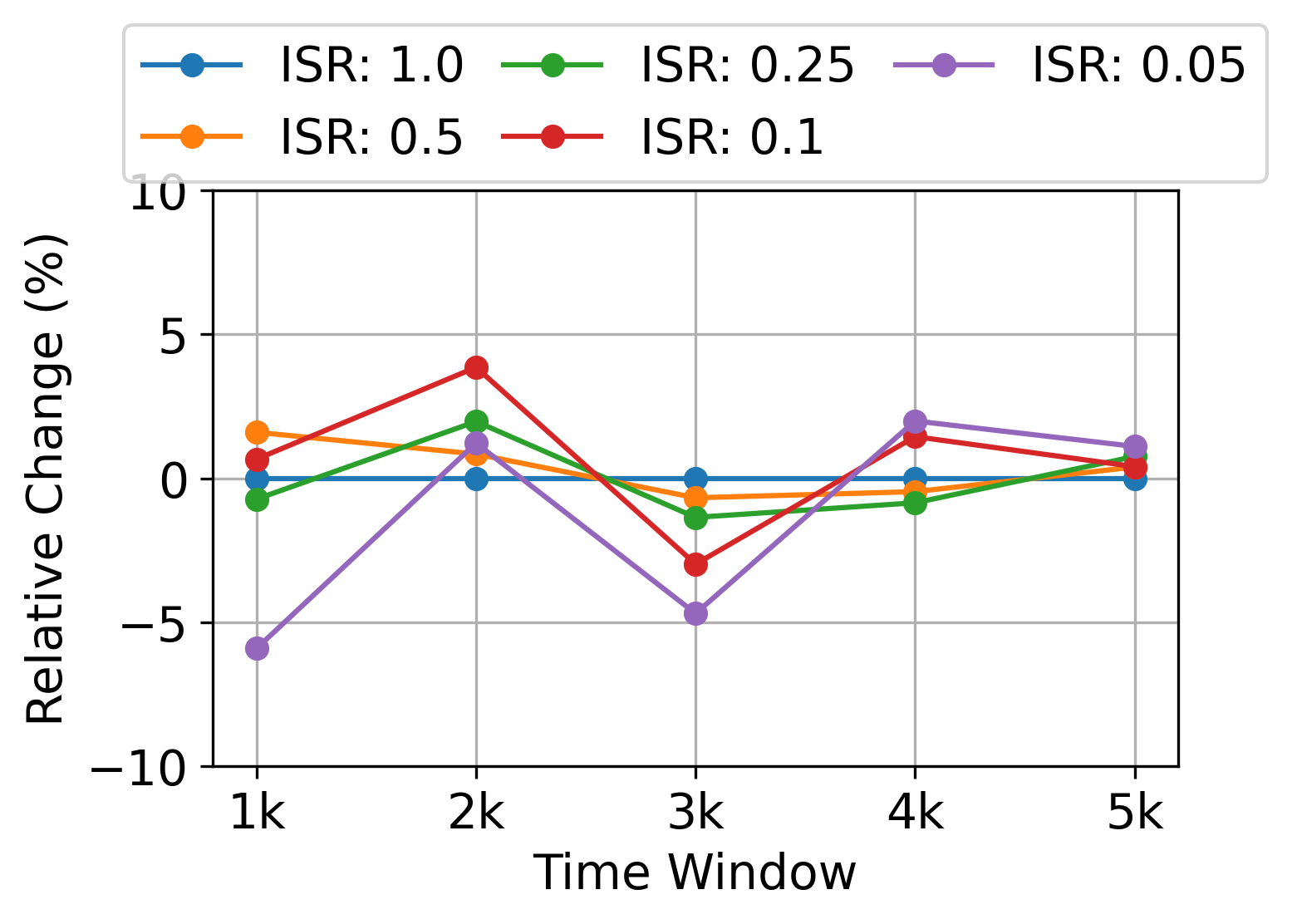}
    \vspace{-2mm}
    \caption{Relative change rate of entropy under different iteration sampling rates.}
    \label{fig:relative-entropy}
\end{figure}

With $\beta$ fixed at 0.25, we then varied $\alpha$.
Figure~\ref{fig:relative-entropy} reports the relative change rate (RCR)
of average entropy per 1,000-iteration window, using $\alpha=1$ as the
baseline. An overly small $\alpha$, such as 0.05, produces an RCR above
5\%, whereas $\alpha=0.1$ yields stable estimates with little deviation.

\begin{table}[t]
\centering
\caption{Time cost under different GSRs ($\beta$)}
\vspace{-3mm}
\resizebox{0.9\columnwidth}{!}{
\begin{tabular}{lcccc}
\toprule
 & \textbf{$\beta$=1.0} & \textbf{$\beta$=0.5} & \textbf{$\beta$=0.25} & \textbf{$\beta$=0.05} \\ \midrule
Calculation Time (ms) & 78.14 & 58.7 & 46.85 & 39.93 \\
\bottomrule
\end{tabular}
}
\label{tab:time_cost_diff_samp_rate}
\vspace{-3mm}
\end{table}

Table~\ref{tab:time_cost_diff_samp_rate} shows that setting $\beta=0.25$ reduces the
per-iteration entropy-computation time by 40\% relative to using all
gradients. Combined with $\alpha=0.1$, the total entropy-estimation time
within each window decreases by approximately 94\%.

\subsection{Analysis of CQM}

Using GPT-2 345M, we compare CQM with three fixed-rank low-rank methods
(64, 32, and 16) and a no-compression baseline. All models are trained
under the same conditions, and perplexity (PPL) is evaluated on the
validation set every 1,000 iterations. Rank 64 is the largest rank
$r_{\max}$ for which compression begins to reduce communication time,
whereas rank 16 is the minimum permitted rank $r_{\min}$. CQM selects
ranks dynamically within this interval.

\begin{figure}[tb]
\centering
\includegraphics[width=\linewidth]{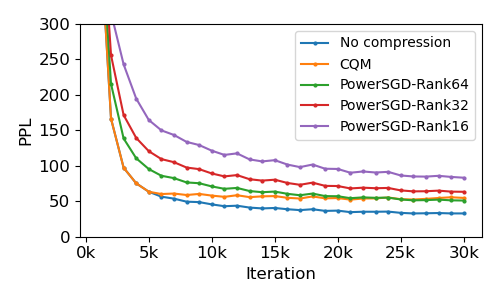}
\vspace{-8mm}
\caption{PPL under different compression strategies.}
\label{fig:ppl}
\vspace{-4mm}
\end{figure}

As shown in Figure~\ref{fig:ppl}, smaller fixed ranks cause greater
information loss. Rank 16 produces substantially higher PPL throughout
training, whereas rank 64 preserves accuracy but provides little
communication benefit.

\begin{table}[t]
    \centering
    \caption{Communication time over 30,000 training steps}
    \label{tab:comm_time_horizontal}
    \vspace{-3mm}
    \resizebox{1.0\columnwidth}{!}{
    \begin{tabular}{lccccc}
        \toprule
        Method & No Compression & Rank=64 & Rank=32 & Rank=16 & CQM (Ours) \\
        \midrule
        Time (h) & 3.0417 & 3.0167 & 1.4833 & 0.7417 & 1.8750 \\
        \bottomrule
    \end{tabular}
    }
    \vspace{-3mm}
\end{table}

\begin{table}[t]
\centering
\caption{Performance metrics (CC, MSE) for varying time window sizes($w$)}
\vspace{-1mm}
\label{tab:performance_diff_window_size}
\resizebox{\columnwidth}{!}{
\begin{tabular}{c|c|ccccc}
\toprule
\textbf{Metric} & \textbf{Model} & \textbf{w=1} & \textbf{w=100} & \textbf{w=500} & \textbf{w=1000} & \textbf{w=2500} \\
\midrule
\multirow{2}{*}{CC} 
  & BERT   & 1.0000 & 0.9891 & 0.9838 & 0.9433 & 0.8004 \\
  & GPT-2  & 1.0000 & 0.9979 & 0.9920 & 0.9807 & 0.9634 \\
\midrule
\multirow{2}{*}{MSE}
  & BERT   & 0.0000 & 0.0059 & 0.0534 & 0.2742 & 1.2181 \\
  & GPT-2  & 0.0000 & 0.0066 & 0.0288 & 0.1561 & 0.8750 \\
\bottomrule
\end{tabular}
}
\vspace{-5mm}
\end{table}

\begin{figure}[b]
\centering
\includegraphics[width=\linewidth]{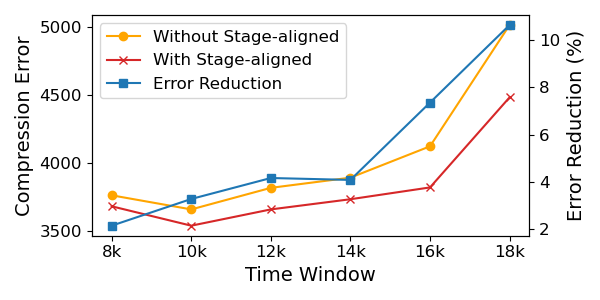}
\vspace{-7mm}
\caption{Effect of stage-aligned rank adaptation on compression error.}
\label{fig:stage-aligned-error}
\vspace{-2mm}
\end{figure}

CQM benefits from full-precision updates during the adaptive warm-up
phase and subsequently adjusts the rank based on gradient dynamics. It
approaches the PPL of rank 64 while requiring substantially less
communication time, as shown in Table~\ref{tab:comm_time_horizontal}.

\subsection{Analysis of DAC}

To evaluate the effect of window size on adaptive rank adjustment, we
measure the fidelity of entropy dynamics under different update
intervals using BERT-345M and GPT-2 345M. Rank adaptation occurs at the
end of each $w$-iteration window. We use the fine-grained schedule
$w=1$, in which the rank is updated every iteration, as the temporal
fidelity baseline.

We quantify similarity to the $w=1$ trajectory using the correlation
coefficient (CC) and mean squared error (MSE).
Table~\ref{tab:performance_diff_window_size} shows that $w=1000$ preserves strong
temporal alignment (CC of 0.9433 for BERT and 0.9807 for GPT-2), while
maintaining $\mathrm{MSE}<0.3$ and reducing the frequency of rank
updates. The larger window $w=2500$ introduces more noticeable
distortion, particularly for BERT.

These results indicate that $w=1000$ balances estimation fidelity and
adaptation overhead, and we therefore use it as the default window size.

To evaluate stage-aware alignment, we compare the full DAC method
against an ablation in which all pipeline stages share one globally
synchronized rank. Figure~\ref{fig:stage-aligned-error} shows that DAC
consistently obtains lower compression error. The absolute error of
both methods increases as the rank is progressively reduced, but
stage-aligned adaptation better follows the heterogeneous evolution of
gradient entropy across layers. The relative error reduction of DAC
over the ablated baseline exceeds 10\% by 18,000 iterations.

\end{document}